\documentclass[11pt]{article}
\newif\ifprintready
\printreadytrue
\usepackage[utf8]{inputenc}
\usepackage[T1]{fontenc}
\usepackage{mathpazo}
\usepackage[margin=1in]{geometry}
\usepackage{microtype}
\usepackage{amsmath,amssymb}
\usepackage{booktabs,multirow,makecell}
\usepackage{graphicx}
\usepackage{tikz}
\usetikzlibrary{positioning,arrows.meta,fit,calc,shapes.geometric}
\usepackage{wrapfig}
\usepackage{fontawesome5}
\usepackage{xcolor}
\usepackage{xspace}
\usepackage{caption}
\usepackage[round,authoryear]{natbib}
\usepackage[colorlinks=true,linkcolor=blue!50!black,citecolor=blue!50!black,urlcolor=blue!50!black]{hyperref}
\usepackage[capitalise,noabbrev]{cleveref}
\usepackage[most]{tcolorbox}
\newif\ifvok
\newenvironment{verified}[1]{\ifcsname vok@#1\endcsname\voktrue\else\vokfalse\fi\ifprintready\vokfalse\fi
  \ifvok\begin{tcolorbox}[enhanced,breakable,colback=white,colframe=blue!55!black,boxrule=0.8pt,arc=1pt,
    left=4pt,right=4pt,top=3pt,bottom=3pt]\fi}
  {\ifvok\end{tcolorbox}\fi}
\InputIfFileExists{verified_ok.tex}{}{}
\newcommand{\chsm}{CHSM8\xspace}
\newcommand{\pt}{\textsc{chsm8-pt}\xspace}
\newcommand{\sft}{\textsc{chsm8-sft}\xspace}
\newcommand{\dpo}{\textsc{chsm8-dpo}\xspace}

\newcommand{\hf}[1]{\href{https://huggingface.co/LegumMagister/#1}{\texttt{#1}}}

\newcommand{\pq}{\emph{``trade queens as early as possible''}\xspace}
\newcommand{\pp}{\emph{``push your pawns aggressively up the board and grab space''}\xspace}
\newcommand{\pc}{\emph{``castle queenside and storm the enemy king with your pawns''}\xspace}
\newcommand{\pa}{\emph{``attack the king at all costs and sacrifice material for the initiative''}\xspace}

\newcommand{\todo}[1]{}
\renewcommand{\todo}[1]{\textcolor{red}{[TODO: #1]}}
\newcommand{\hfd}[1]{\href{https://huggingface.co/datasets/LegumMagister/#1}{\texttt{#1}}}

\newcommand{\rankmark}[3]{\tikz[baseline=-0.6ex]\node[regular polygon,regular polygon sides=3,shape border rotate=#2,
  fill=#1,text=white,inner sep=0pt,minimum size=10.5pt,font=\fontsize{5.5}{5.5}\selectfont\bfseries]
  {\raisebox{\ifdim#2pt=0pt -0.45ex\else 0.25ex\fi}{#3}};}
\newcommand{\rankup}[1]{\rankmark{green!55!black}{0}{#1}}
\newcommand{\rankdown}[1]{\rankmark{red!75!black}{180}{#1}}
\newcommand{\rankeq}{\tikz[baseline=-0.6ex]\node[inner sep=0pt,minimum size=10.5pt,text=gray,font=\scriptsize]{=};}
\newcommand{\grpsup}{{\color{black!75}\faChessPawn}}
\newcommand{\grpadv}{{\color{orange!85!black}\faRocket}}
\newcommand{\grpsearch}{{\color{blue!60!black}\faSearch}}

\usepackage{float}
\usepackage{needspace}
\usepackage{etoolbox}
\pretocmd{\section}{\clearpage}{}{}

\usepackage{pgfplots}
\pgfplotsset{compat=1.17}

\title{\ifprintready\else{\small\sffamily\color{blue!55!black}Work in progress: blue frames mark verified text}\\[10pt]\fi
\chsm: A Chess Transformer that Follows Natural-Language Style Prompts\\[2pt]
\large Technical report}
\author{Ilia Badanin}
\date{\vspace{-8pt}October 2026}

\begin{document}
\maketitle
\begin{abstract}
\begin{verified}{abstract}
We present \chsm, a 210M-parameter chess transformer that plays moves conditioned on a free-text description of a playing
style, e.g.\ \emph{``go for the Sicilian Defense''}, \emph{``solid, safe, risk-free chess''} or
\emph{``You are Magnus Carlsen in his prime''}.
The model is pretrained without search on 1.07B filtered Lichess games, using a factorized move representation and
legal-move masking.
It reaches 1837 Elo against a strength-limited Stockfish and, in 200-game matches, beat every other public model we
tested that was likewise trained without preference tuning, rewards or search, including one with 13 times as many
parameters.
We then fine-tune it to follow text on a synthetic dataset: Gemma-4-31B describes the playing style of 3M game sides,
a 1B Gemma-3 model distilled from the 2.4M of those descriptions that contradict nothing on the board labels the rest,
and every one of 30M selected game sides gets five descriptions.
Supervised fine-tuning attaches a frozen text encoder through gated cross-attention, and direct preference optimization
(DPO) uses the same moves under a mismatched description as the rejected sample.
After DPO, style adherence rises on every prompt we test (for example, queens traded early 53\%$\to$67\%, pawn
moves 29\%$\to$32\% of all moves, queenside castling 97.3\%$\to$99.7\%), and strength without a prompt stays within noise of SFT (1792 vs.\ 1797 Elo).
Design choices at every stage were made by fixed-budget automated search loops on top of standard architectures and
training recipes.
We document the full pipeline and the decisions behind it, report the ablations that failed, and evaluate the models
across many style prompts and kinds of games, with 1,000 games per setting.
The work is fully open source and reproducible: weights, datasets with all training labels, prompts,
configurations, inference code and playable demos are public and free to build on.
\end{verified}
\end{abstract}

\ifprintready\vspace*{\fill}\fi
\begin{figure}[H]
\ifprintready\else\vspace{-6pt}\fi\centering
\scalebox{\ifprintready 1.2\else 0.85\fi}{\begin{tikzpicture}[
  font=\small, >={Stealth[length=2mm]},
  box/.style={draw, rounded corners=3pt, align=center, inner sep=5pt},
  io/.style={draw=gray!50, rounded corners=2pt, align=center, inner sep=4pt, text width=46mm}]
\node[io] (prompt) {\textbf{style prompt}\\\emph{``castle queenside and storm the king''}};
\node[io, below=5mm of prompt] (hist) {\textbf{game so far}\\\texttt{1.\,e4 c5 2.\,Nf3 d6 3.\,d4 \ldots}};
\node[box, fill=orange!12, text width=30mm, minimum height=24mm, right=10mm of $(prompt.east)!0.5!(hist.east)$] (model)
  {\textbf{\chsm}\\[2pt]chess transformer\\conditioned on\\the prompt};
\node[io, text width=30mm, right=10mm of model] (move) {\textbf{next move}\\always legal,\\in the asked style};
\draw[->] (prompt.east) -- (prompt.east -| model.west);
\draw[->] (hist.east) -- (hist.east -| model.west);
\draw[->] (model) -- (move);
\end{tikzpicture}}
\\[2pt]{\small With this prompt, queenside castling: 13\% $\rightarrow$ 94\% of games against Stockfish.}
\label{fig:teaser}
\end{figure}
\ifprintready\vspace*{\fill}\fi
\newpage
\tableofcontents
\newpage
\section{Introduction}
\begin{figure}[H]
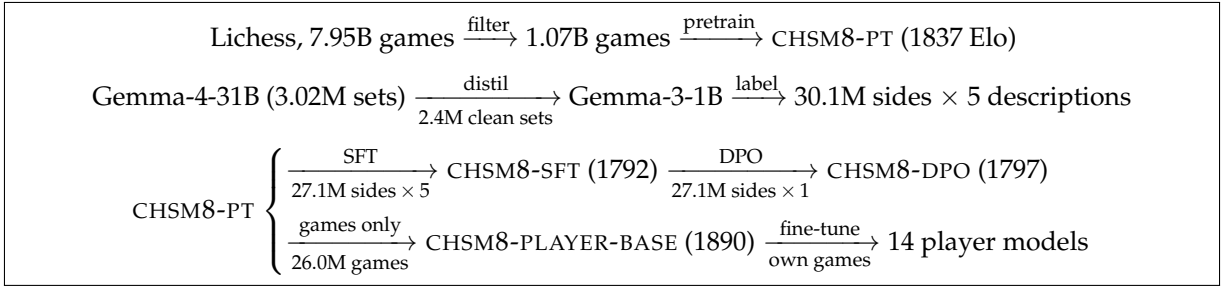

\centering\small
\fbox{\parbox{0.96\linewidth}{\centering
Lichess, 7.95B games $\xrightarrow{\text{filter}}$ 1.07B games $\xrightarrow{\text{pretrain}}$ \pt (1837 Elo)\\[6pt]
Gemma-4-31B (3.02M sets) $\xrightarrow[\text{2.4M clean sets}]{\text{distil}}$ Gemma-3-1B
$\xrightarrow{\text{label}}$ 30.1M sides $\times$ 5 descriptions\\[6pt]
\pt\ $\left\{\begin{array}{@{}l@{}}
\xrightarrow[\text{27.1M sides}\,\times\,5]{\text{SFT}} \text{\sft\ (1792)}
\xrightarrow[\text{27.1M sides}\,\times\,1]{\text{DPO}} \text{\dpo\ (1797)}\\[10pt]
\xrightarrow[\text{26.0M games}]{\text{games only}} \text{\textsc{chsm8-player-base} (1890)}
\xrightarrow[\text{own games}]{\text{fine-tune}} \text{14 player models}
\end{array}\right.$}}
\caption{Pipeline overview. From \pt the work forks: the style branch (SFT, then DPO) and, in parallel, a games-only
branch for player models. Below each arrow is the training data; Elo (no prompt) is in brackets, measured with one
protocol (\cref{sec:eval}). The main DPO model saw 0.56M of the 27.1M training sides (\cref{sec:dpo}).}
\label{fig:pipeline}
\end{figure}

\begin{verified}{intro}
Chess remains a hard problem for neural networks without search.
Transformers trained on engine annotations or on human games now play at a useful level directly from the board or the
move history \citep{ruoss2024amortized,mcilroy2020maia,tang2024maia2}, but they still make tactical errors that a
search engine would not.
None of them, however, lets a user simply \emph{say} how the model should play.
That is not a matter of strength but of experience: a player who wants a sparring partner that attacks, trades queens
early or plays like a favorite grandmaster has no way to ask for it.
Rating-conditioned models \citep{tang2024maia2,zhang2025allie} control strength, and player-specific fine-tunes control
identity, but neither accepts an open-ended request such as \pc.

\chsm is built for this kind of use, aimed at user experience and entertainment rather than grandmaster-level play.
Its models nevertheless stay strong without search, so they are a sound base for others to build on.
For new work we recommend \textsc{chsm8-player-base}: \pt continued on high-quality games (players rated $\ge2200$),
a plain decoder with self-attention only, no cross-attention, and the strongest of our models (1890 Elo).
It is trained in a staged pipeline (\cref{fig:pipeline}):
\begin{enumerate}\setlength\itemsep{0pt}
\item \textbf{Pretraining} a move model on 1.07B games, with factorized move tokens, a soft illegal-move penalty in
training and legal-move masking at decoding
(\cref{sec:pretraining}).
\item \textbf{Labelling} 30M game sides with natural-language style descriptions by a small distilled LLM that is
grounded in measured game features (\cref{sec:labeler}).
\item \textbf{Supervised fine-tuning} with gated cross-attention to a frozen text encoder (\cref{sec:sft}).
\item \textbf{Preference optimization} with label-swap pairs, which needs neither sampling nor a reward model
(\cref{sec:dpo}).
\item \textbf{Inference}: the model scores every legal move from its four factorized heads and plays the best one, with
no search (\cref{sec:compose}); demos run it in the browser or on a local GPU (\cref{app:artefacts}).
\item \textbf{Player models}, a separate branch from \pt: games-only continued training into \hf{chsm8-player-base},
then 14 fine-tunes that imitate individual top players from their own online games (\cref{sec:pt-results}).
\end{enumerate}
\Cref{sec:eval} describes how each stage is evaluated, and \cref{sec:results} gives the final comparison.

\paragraph{Method.}
We deliberately build on established practice rather than inventing new components.
The decoder follows the Llama recipe \citep{touvron2023llama}; schedules follow GPT-3 and OLMo~2
\citep{brown2020gpt3,olmo2}; text conditioning uses Flamingo-style gated cross-attention \citep{alayrac2022flamingo};
labels are distilled from a larger teacher; and preferences are optimized with DPO \citep{rafailov2023dpo}.
Within this standard frame, the design choices at every stage were made by an automated search in the style of
\texttt{autoresearch} \citep{karpathy2026autoresearch}: a queue of configurations, each trained for a fixed wall-clock
budget, scored on a fixed validation set and logged with all its metrics, with the next round planned from the results.
We ran such loops for the pretraining architecture, the labeler, the SFT recipe and the DPO recipe
(\cref{fig:pt-search,fig:labeler-search,fig:sft-search,fig:dpo-search}).
Every run and its configuration are released with the code.

\paragraph{Contributions.}
\begin{enumerate}\setlength\itemsep{0pt}
\item A fully open and reproducible pipeline: weights, datasets (including the teacher labels used to train both the
labeler and the style model), the complete recipe with prompts and configurations, inference code and playable demos are
public (\cref{app:artefacts}); the training code and search logs will follow.
\item To our knowledge, the first chess model that plays by a free-form natural-language description of a style, rather
than by a rating, a player identity or a templated prompt (\cref{sec:sft,sec:dpo}; closest work in \cref{sec:related}).
Its 210M-parameter base, without search, beat every other public model we tested that was likewise trained without
preference tuning, rewards or search, from 23M to 2.8B parameters (\cref{tab:versus}).
\item New datasets that did not exist before (\cref{sec:data}): a high-quality corpus of 1.07B games, filtered from the
7.95B-game Lichess archive by rating, time control and game quality; the online games of the FIDE top 100 crawled from
Chess.com and Lichess, including the games of the 14 players we fine-tuned on; five style descriptions for each of
30.1M game sides; and the teacher labels and distillation set behind the labeler.
\end{enumerate}
\end{verified}

\section{Related work}\label{sec:related}
\begin{verified}{related}
Our topic is \emph{style}: making a chess model play the way a free-text description asks.
Most chess research instead targets strength, human-likeness or reasoning, and we summarize the strongest recent
results in each before turning to the work closest to ours.

\paragraph{Strength: search and searchless play.}
AlphaZero \citep{silver2018alphazero}, Leela Chess Zero \citep{lc0transformer2024} and Stockfish \citep{stockfish}
combine a learned or hand-tuned evaluation with deep search and remain far stronger than any model without search.
Lc0's transformer networks raised raw policy strength by about 300 Elo over its best convolutional net
\citep{lc0transformer2024}.
Without search, a 270M transformer trained on Stockfish action values reaches a Lichess blitz Elo of 2895
\citep{ruoss2024amortized}.
ChessLFM \citep{labonne2026chesslfm} fine-tunes a 230M general language model (LFM2.5) by Stockfish distillation, with
one token per move; it predicts its own win probability before the move and reaches about 2000 Elo with a shallow
depth-3 search on top.
MAV \citep{schultz2025mav} trains one transformer on several board games and reaches grandmaster level in chess with
either external search (MCTS guided by the model, without an engine) or internal search written out in context.
Our model plays without search and is far weaker than these systems; strength is not its goal.

\paragraph{Human-like and individual play.}
Maia \citep{mcilroy2020maia} and Maia-2 \citep{tang2024maia2} predict human moves at a given rating.
Maia-3, built on the Chessformer architecture \citep{monroe2026chessformer}, is the current state of the art in human move
prediction (57.1\% move matching with 79M parameters), and Matilda \citep{carlson2026matilda} adds engine search to a frozen
human policy to predict the moves of very strong players.
Otter \citep{kumar2026otter} conditions on move history and the clock, Allie \citep{zhang2025allie} adds a little search
to a human-aligned model, and \citet{zhang2024transcendence} show that low-temperature decoding lets imitation models
exceed the players they learn from.
Maia4All \citep{tang2025maia4all} models individual players from as few as 20 games.
Character-level PGN models such as Chess-GPT \citep{karvonen2024chessgpt} learn chess from raw game text.
We follow the imitation line, but our control signal is free text rather than a rating or a player identity, and we
compare against several of these models head to head (\cref{sec:pt-results}, \cref{tab:versus}).
Our own player models follow the fine-tuning route (\cref{sec:pt-results}).

\paragraph{Language models that reason about chess.}
General-purpose LLMs remain weak players: in ChessArena \citep{liu2025chessarena} no tested LLM beat Maia-1100, and the
Kaggle Game Arena \citep{kaggle2025gamearena} now ranks frontier LLMs by playing them against each other.
Specialized models close much of the gap.
C1 \citep{tang2026c1}, a 4B model trained by distilling expert explanations and then by reinforcement learning, solves
48.1\% of its chess problems, ahead of most open and many proprietary models.
Queen \citep{bhaskar2026queen} connects a chess encoder to an instruction-tuned LM through cross-attention, reaches
2697 Elo and explains its moves.
ChessGPT \citep{feng2023chessgpt} trains language models on mixed chess and text data and learns a CLIP-style
game--language embedding.
These models put language on the \emph{output} side, to explain or justify moves; we put it on the \emph{input} side, to
steer how the moves are chosen.

\paragraph{Language-steered chess policies.}
The closest work is UniMaia \citep{siu2026unimaia}, which steers a frozen Lc0-based policy with a parameter-efficient
text encoder and ControlNet-style conditioning, trained on metadata-augmented Lichess games with semi-automatically
generated prompts, and controls aspects such as the opening and playing strength.
\chsm differs in three ways: its prompts are open-ended style descriptions written by an LLM from features measured
on the board, rather than templates over game metadata (\cref{sec:labeler}); the whole move model is trained, with text entering through
gated cross-attention \citep{vaswani2017attention,alayrac2022flamingo} (\cref{sec:sft}); and adherence is improved with
preference training (\cref{sec:dpo}) and measured by how the model actually plays under each prompt
(\cref{sec:eval}, \cref{fig:style-heatmap}).

\paragraph{Preference optimization and guidance.}
DPO \citep{rafailov2023dpo} optimizes a Bradley--Terry \citep{bradley1952rank} preference objective directly against a
frozen reference; IPO \citep{azar2024ipo} replaces the logistic loss with a squared one.
GRPO \citep{shao2024grpo} normalizes rewards within groups of samples.
Classifier-free guidance \citep{ho2022cfg} strengthens conditioning at inference time.
Our DPO pairs are not sampled from the model: they reuse the same human game under a matched and a mismatched
description (\cref{sec:dpo}).

\paragraph{Labelling with LLMs and distillation.}
We use LLMs as annotators and distil \citep{hinton2015distilling} the best teacher, Gemma-4-31B, into Gemma-3-1B
\citep{gemma3}, served with vLLM \citep{kwon2023vllm}.
Students compared include T5Gemma \citep{zhang2025t5gemma}, Qwen3 \citep{qwen3}, FLAN-T5 \citep{chung2022flan} and the
diffusion model LLaDA \citep{nie2025llada}; the second teacher was Apertus-v1.5-70B \citep{apertus2025} (\cref{tab:labeler,fig:labeler-search,tab:labelsrc}).
Text encoders compared for conditioning include EmbeddingGemma \citep{vera2025embeddinggemma}, Qwen3-Embedding
\citep{zhang2025qwen3emb}, BGE \citep{xiao2023bge} and MiniLM \citep{wang2020minilm,reimers2019sbert} (\cref{fig:sft-enc}).
\end{verified}

\section{Evaluation protocol}\label{sec:eval}
\begin{verified}{evaluation}
We evaluate on several levels, each with its own measurements:
\begin{enumerate}\setlength\itemsep{0pt}
\item \textbf{strength} of every model against a fixed engine, without a prompt (below; \cref{fig:pt-elo,tab:headline});
\item \textbf{head-to-head} matches of \pt against other public chess models (\cref{tab:versus});
\item \textbf{style adherence in play}: how the model's own moves change under a prompt (below;
\cref{tab:headline,fig:style-heatmap});
\item \textbf{likelihood}: whether the right description makes the real moves more likely than a wrong one
(\cref{tab:valce});
\item \textbf{label quality}: whether descriptions agree with the board, and how much signal each label source carries
(\cref{tab:labeler,tab:labelsrc});
\item \textbf{player models}: move matching on each player's held-out games (\cref{sec:pt-results});
\item \textbf{noise}: how many games each comparison needs, and how we select among runs that are within noise
(\cref{sec:noise,sec:z}).
\end{enumerate}

\paragraph{Playing strength.}
The model plays greedy legal moves (the highest-scoring legal move, no sampling, no search) against Stockfish~17.1
\citep{stockfish} limited to \texttt{UCI\_Elo}~2000, with colors alternating and games capped at 240 plies.
Unless a prompt is named, Elo is measured \emph{without} a prompt.
From the score $S=(W+D/2)/N$ we report $\text{Elo}=2000+400\log_{10}\!\big(S/(1-S)\big)$ \citep{elo1978rating} with a
95\% interval from the per-game standard error.
The \texttt{UCI\_Elo} scale is a reproducible engine calibration, not a FIDE or Lichess rating.

Legality is not something we measure.
Training uses a cheap legality mask or penalty on each factorized head (\cref{sec:tokens}), but at inference a chess
library tracks the board and only legal moves are scored: python-chess \citep{pythonchess} in our evaluation, and the
Rust library \texttt{shakmaty} in the demos.
Every generated move is therefore legal by construction.

\paragraph{Why each stage costs some strength.}
We expect strength without a prompt to fall slightly at each stage, and it does (\cref{fig:pipeline}).
Pretraining imitates players rated 1800 and above.
SFT then optimizes a different objective, matching a description rather than playing the best move, on games chosen for
their descriptions rather than their quality.
It also adds a new input: pretraining had no text, so its cross-attention blocks were left untrained, and SFT has to
learn them from scratch while the decoder adapts (\cref{sec:sft-final}).
DPO, finally, only has to keep strength where SFT left it.

\paragraph{Style adherence.}
For text-conditioned models we play under fixed prompts and measure behavior on the model's own moves.
The four main prompts, each with a metric that should rise when the prompt is followed, are:
\begin{itemize}\setlength\itemsep{0pt}
\item \textbf{queens}: share of games with queens traded by ply 40, under \pq;
\item \textbf{pawns}: pawn moves per own move, under \pp;
\item \textbf{queenside}: queenside share of castling, under \pc;
\item \textbf{attack gap}: Elo under \pa minus Elo with no prompt.
\end{itemize}
\Cref{fig:style-heatmap} extends these to eight behaviors per prompt.
A broader 31-prompt evaluation (strong, weak and style prompts) was used for SFT (\cref{sec:sft-search}).

The attack gap is not a reliable metric, and we treat it as a qualitative signal.
A model that takes ``attack at all costs and sacrifice material'' literally gives up material without a plan and,
without search, loses to an accurate engine.
A large negative gap with unchanged no-prompt strength therefore suggests the prompt is followed faithfully, not that the
model got weaker; but it cannot separate faithful attacking play from simply playing badly.

\paragraph{Style by self-play.}
For style we also let the prompted model play \textsc{chsm8-player-base} without text, with all games advancing one move
at a time as a single GPU batch.
On one 4-GPU node this plays 12,000 games in 10 minutes, about five times faster in wall-clock time than the same number
of games against Stockfish, which is limited by one CPU thread per game.
The opponent is itself only trained on human games, so it does not bias the style metrics.
Both sides play their most likely move, so games with the same prompt would repeat; each book opening is therefore
played twice with colors swapped, and only the prompted side's moves after the book are measured.
Every game can carry its own prompt, which makes a baseline of random descriptions from the test split as cheap as one
fixed prompt.

\paragraph{Likelihood metrics.}
For conditioned models we also report cross-entropy (nats per move, summed over the factorized heads) on the described
side's moves with the right description, a wrong description (another game's) and no description.
The gap between right and wrong measures how much the text is used.

\paragraph{Noise.}
Elo intervals are about $\pm$70 at 100 games, $\pm$50 at 200, $\pm$30 at 1,000 and $\pm$21 at 2,000 games.
The attack gap is a difference of two Elo estimates and is about 1.4$\times$ noisier.
\Cref{sec:noise} shows that this noise, not model differences, limited the final DPO comparisons.
\end{verified}

\section{Data}\label{sec:data}
\begin{verified}{data-1}
\subsection{Pretraining corpus (\hfd{chsm8-pre-training})}
We start from all 162 monthly \href{https://database.lichess.org}{Lichess} standard-rated dumps \citep{lichessdb}, 2013-01 to 2026-06: 7.95B games and
2.51\,TB compressed.
A game is kept only if it passes all of the following filters:
\begin{itemize}\setlength\itemsep{0pt}
\item \textbf{speed}: estimated duration $\text{base}+40\times\text{increment}\ge180$\,s (blitz, rapid, classical);
\item \textbf{rated}, both players \textbf{human} (no \texttt{BOT} title), \textbf{standard} variant;
\item \textbf{finished}: result 1-0, 0-1 or draw; time forfeits, abandonments and rule infractions dropped (24.5\% of
blitz games end on time);
\item \textbf{strength}: both players rated $\ge1800$;
\item \textbf{parse}: every move replays legally with python-chess \citep{pythonchess} (0 failures in 267M games
checked).
\end{itemize}
Each kept game is cut at the first ply where the side to move has less than 15\,s on the clock.
This removes time scrambles while keeping 94\% of blitz plies and 99\% of rapid and classical plies.
Games before 2017-04 carry no clock data and are kept whole (2\% of the archive).
The result is 1,068,263,794 games (12.4\% of the raw archive) with 71.1B moves, 66.5 plies per game on average.
Results lean toward White: 49.1\% of games end 1-0 and 44.7\% 0-1, with 6.2\% draws.

\paragraph{Processing.}
The corpus is built in tiers, so that a changed threshold never requires rescanning the 2.51\,TB archive
(\cref{tab:dataproc}).
The scan (tier 2a) reads every PGN once, replays every move and applies only the cheap filters (speed, rated, human,
finished); everything that might be revisited, such as ratings, clocks and termination, is stored as a column.
Tier 2b applies the strength filter, drops time forfeits and cuts time scrambles in minutes.
Tier 3 converts every game into factorized move tokens (\cref{sec:tokens}) plus a legal-move mask per position, and
tier 3p packs them into training rows.

\end{verified}
\begin{table}[H]
\centering\footnotesize\setlength\tabcolsep{4pt}
\begin{tabular}{@{}llrll@{}}
\toprule
tier & conversion & output & process time & wall-clock time \\
\midrule
1 & download, 162 monthly \texttt{.pgn.zst} & 2.51\,TB & 6.5\,h & 6.5\,h \\
2a & PGN $\rightarrow$ 16-bit move ids + metadata (cheap filters) & 784\,GB & 170\,h & \makecell[l]{22\,h (8 months at a time,\\16 workers each); largest\\month 1.9\,h} \\
2b & strength, time forfeits, clock cut & 156\,GB & 4.7\,h & \makecell[l]{4.7\,h (one month at a time);\\largest month 7\,min} \\
3 & move ids $\rightarrow$ factorized tokens + legal masks & 1.39\,TB & 76\,h & \makecell[l]{about 1.2\,h (96 workers);\\largest month 1.2\,h} \\
3p & greedy packing into 512-position rows & 1.4\,TB & --- & 12--13\,min (162 workers) \\
\bottomrule
\end{tabular}
\caption{Data processing tiers, from the raw archive to training rows, as measured in our build. Process time is the
sum over all monthly processes; wall-clock time is the real time of the build at the parallelism shown. Each month is
an independent process, so at full parallelism the largest month sets the wall-clock time. The released conversion
scripts, rewritten for the public dataset, rebuild training rows from its Parquet files in about 2\,h on one node
(dataset card).}
\label{tab:dataproc}
\end{table}

\begin{figure}[H]
\centering
\includegraphics[width=0.6\linewidth]{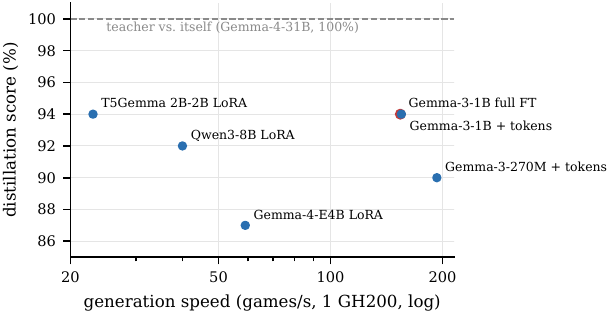}
\caption{Labeler search: distillation score against generation speed for the students of \cref{tab:labeler}. Speeds are vLLM on one GH200, except T5Gemma (Hugging Face generation) and Qwen3-8B and Gemma-4-E4B (vLLM speed proxy with base weights).}
\label{fig:labeler-search}
\end{figure}

\begin{verified}{data-2}

Memory, not compute, was the binding constraint.
The first packing run buffered about 7\,GB per worker, 1.5\,TB over 200 workers on a node with 856\,GB, and was killed;
workers now stream their rows to disk.
A second version copied the 17\,GB game index into every worker before starting; each worker now opens only its own
slice.
For SFT, the 77\,GB array of description embeddings is filled in RAM and written once, because row-by-row random writes
to the shared filesystem stalled.

\paragraph{An era trap.}
Speed must be decided by arithmetic, not by the PGN \texttt{Event} label.
Before 2017-04 Lichess had no rapid category, so the label ``Classical'' covered games of 8 minutes and up; afterwards it
covers only 25 minutes and up.
A label-based filter silently kept 28.8\% of 2013-01 games but only 0.6\% of 2026-06 games, a 52$\times$ difference from
identical code.
The arithmetic test above means the same thing in every era.
The share of games that survive our filters also varies by month; it peaks at 27.5\% around 2021-10 to 2021-12,
against 20--23\% in neighboring months.
The same months coincide with the loss bump in pretraining (\cref{sec:curricula}); we do not know the cause, and
reshuffling did not remove it.

\paragraph{Packing and data order.}
Games are tokenized once (\cref{sec:tokens}) and greedily packed, without truncation, into rows of 512 positions.
Padding is 7.7\%, and 1\% of games are held out for validation.
The packed corpus is laid out chronologically, and reading it in order would impose an accidental old-to-new curriculum.
We therefore build the training stream (\texttt{3p-s}) as a fixed permutation:
\begin{itemize}\setlength\itemsep{0pt}
\item the first 5\% of positions (the learning-rate warmup) is drawn uniformly from the whole corpus;
\item after that, a uniformly shuffled two-thirds of the data is interleaved with a reverse-chronological one-third,
newest months first;
\item no packed row occurs twice, so the stream is one pass over 78.32B positions.
\end{itemize}
\Cref{sec:pt-run} describes how this order interacted with the training loss.

\subsection{Post-training corpus (\hfd{chsm8-post-training})}
For style training the selection is player-first rather than game-first as in pretraining.
Starting from the 1.07B pretraining games (so both players are already rated $\ge1800$), we take every side whose own
player is rated $\ge2200$ in that game, group these sides by Lichess account, and keep at most 100 per account, drawn
uniformly at random from all of that account's qualifying games rather than the newest.
This gives 30,085,219 sides of 871,497 players; 200,028 players hit the cap, and the median player has 11 sides.
The released set holds 30,083,545 of these sides, from 28.9M unique games.
Splits are by game (27.08M train, 1.50M validation, 1.50M test sides), so both sides of a game land in the same split.
The 27.08M training sides come from 26.0M distinct games, because in about 1.1M games both players qualify.
Style training reads side appearances, so such a game is seen twice, once under each side's description.
The games-only model instead trains on whole games, like pretraining, so duplicates are removed and it reads each of
the 26.0M games once; this is the number in \cref{fig:pipeline}.
The mix is 75\% blitz, 24\% rapid and 1\% classical, and game length (67 plies) matches the pretraining corpus.
Each side carries five style descriptions (\cref{sec:labeler}).

\paragraph{How the descriptions were made.}
The 30.1M sides are a stricter selection from the same Lichess games as the pretraining corpus.
We screened several teacher LLMs first; Gemma-4-31B wrote the most grounded descriptions, Apertus-v1.5-70B carried much
less signal about the game, and the others were rejected for invented events, generic text or failed parsing
(\cref{sec:teachers,sec:label-signal}).
At the teacher's 7.5 sides per second per node, describing the 30.1M selected sides with Gemma-4-31B would take about
1,100 node-hours, and both sides of their 28.9M games about 2,100; either way it was too slow and costly.
The teacher therefore described 3.02M sides, a
1B Gemma-3 student was distilled from the 2.4M teacher sets that contradict nothing on the board, and the student
described the remaining 27.1M sides.
Trained only on these clean sets, the student contradicts the board in 0.5\% of labels against 4.7\% for the teacher
(\cref{tab:labeler}).
Its greedy decoding also removes the teacher's sampling noise, so it agrees with the teacher more consistently than a
second teacher sample does (\cref{sec:labeler}, \cref{fig:labeler-search}).
On one 4-GPU node the teacher labeled about 7.5 sides per second on average (12.5 at best), while the student
labels about 520, roughly 70 times faster.

\subsection{Top players}
We crawled the \href{https://www.chess.com/news/view/published-data-api}{Chess.com} and
\href{https://lichess.org/api}{Lichess} accounts of the \href{https://ratings.fide.com/top_lists.phtml}{FIDE top 100}
(\hfd{chsm8-top-players}, \cref{app:artefacts}) as of October 2026.
For each player we keep up to 10k recent games, preferring slower time controls.
99 players have online games (one has none findable), with 25 to 10,000 games each.
That gives 602,357 records, one per top player's side, from 561,292 games: in 41,065 games both players are in the top
100, so the game appears once for each side.
542,109 records are for training; the 60,248 others form the validation and test splits.
Each record is paired with a \emph{``play like \{player\}''} prompt drawn from 49 templates and name aliases.
Four historical players (Fischer, Kasparov, Karpov, Polgar), taken from \href{https://www.pgnmentor.com/files.html}{PGN Mentor}, are used only for player-specific
fine-tunes.

\subsection{Data for each training stage}\label{sec:curricula}
Every stage reads a different slice of these corpora, in a different order (\cref{tab:curricula}).

\end{verified}
\begin{table}[H]
\centering\scriptsize\setlength\tabcolsep{4pt}
\begin{tabular}{@{}p{0.2\linewidth}p{0.36\linewidth}p{0.38\linewidth}@{}}
\toprule
stage & data & order \\
\midrule
\makecell[tl]{pretraining\\\hf{chsm8-pt}} & 1.07B games, 78.32B positions, one pass\newline \hfd{chsm8-pre-training} & 5\% uniform, then 2/3 shuffled
mixed with 1/3 newest-to-oldest; from 60\% to the end globally reshuffled  \\
\midrule[0.2pt]
\makecell[tl]{games-only\\\hf{chsm8-player-base}} & 26.0M games of the post-training selection (train split), no text, one pass\newline \hfd{chsm8-post-training} & uniformly shuffled  \\
\midrule[0.2pt]
\makecell[tl]{player fine-tunes\\\texttt{chsm8-player-*}} & one player's own games, own moves only, 2 passes\newline \hfd{chsm8-top-players} & uniformly shuffled  \\
\midrule[0.2pt]
\makecell[tl]{SFT with text\\\hf{chsm8-sft}} & 27.08M described sides, all five bands, one pass; top-player games with \emph{``play like
\{player\}''} prompts in about 6\% of batches (3 passes over that corpus)\newline \hfd{chsm8-post-training}, \hfd{chsm8-top-players} & uniformly shuffled across time;
length-bucketed batches; 10\% of descriptions dropped  \\
\midrule[0.2pt]
\makecell[tl]{DPO\\\hf{chsm8}} & the same 27.08M sides as label-swap pairs, one random band per game; the main model saw 0.56M, the
full-pass model all\newline \hfd{chsm8-post-training} & uniformly shuffled; length-bucketed batches  \\
\bottomrule
\end{tabular}
\caption{Data and data order for each training stage, with the released model under each stage and the released dataset under its data. The DPO model is
released as \hf{chsm8}; \dpo in this report is the same model.}
\label{tab:curricula}
\end{table}

\begin{figure}[H]
\centering
\includegraphics[width=\linewidth]{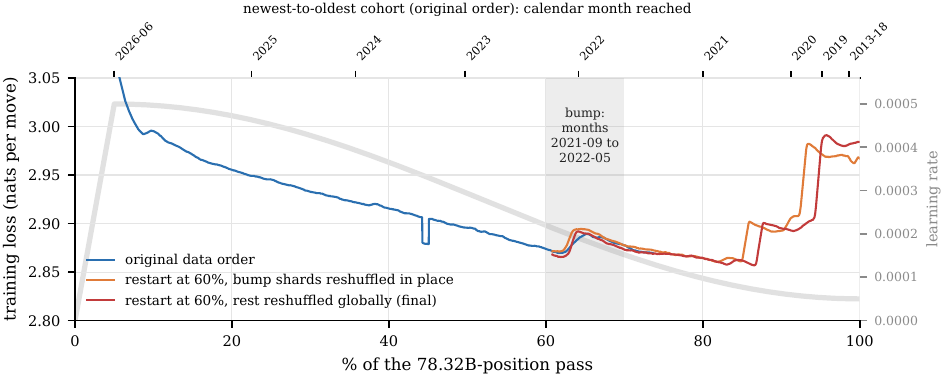}
\caption{Pretraining loss over the pass for every branch (moving average). The shaded band is the 60--70\% bump caused by
the 2021-09 to 2022-05 shards; the late rise from about 85\% appears in both restarted branches.}
\label{fig:pt-loss}
\end{figure}

\begin{verified}{data-3}

\paragraph{Passes over player games.}
The player corpus is small next to the description corpus, so it is repeated.
In SFT, the share of player games mattered and the number of wordings per game did not (\cref{sec:sft-search}): a
flat 6\% of batches over the whole run needs three passes over the 542k player records, each with fresh wording.
The ablations behind both choices are in \cref{sec:sft-search} (SFT) and \cref{sec:pt-results} (player fine-tunes).
The player fine-tunes use two passes at learning rate $3{\times}10^{-5}$, the recipe with the best held-out move match
among those that stayed within 75 Elo of the base; four passes at a lower rate were not better, and LoRA and DPO fits
were worse (\cref{sec:pt-results}).

\paragraph{Pretraining order.}
The packed corpus is chronological, so reading it in order would be an accidental curriculum from old to new chess.
There are two natural choices on purpose: end the pass on the most modern games, so they are learned last, or start
with them at peak learning rate, so they receive the largest updates.
We took a middle path.
The first 5\% of positions, the learning-rate warmup, is drawn uniformly from the whole corpus.
After that, a uniformly shuffled two-thirds of the data is interleaved with a one-third cohort read newest month first,
so that modern games arrive while the learning rate is high but never dominate a stretch of training.

\paragraph{Trouble with the pretraining order.}
The interleaved cohort did not stay harmless (\cref{fig:pt-loss}).
Between 60\% and 70\% of the pass the training loss rose by 1.1\%, on the shards that hold Lichess months 2021-09 to
2022-05, the months with the highest filter survival rate in the corpus.
Restarting from 60\% with those shards reshuffled in place reproduced the bump, so the cause was the games in that
window, not their order inside it.
The final branch restarts at 60\% with all remaining rows globally reshuffled.
Even so, both restarted branches show a second, larger rise from about 85\% of the pass that does not recover
before the end (2.86 to 2.98 nats).
We have not found its cause; strength measured during those last 15\% stays flat (\cref{fig:pt-elo}), and the final model
is the one evaluated throughout.
The two abandoned branches cost about 13\% of a pass in compute.

\end{verified}

\section{Pretraining}\label{sec:pretraining}
\begin{verified}{pt-1}
\subsection{Factorized move tokens}\label{sec:tokens}
A flat vocabulary of move strings depends on the notation (SAN, UCI, LAN) and forces the model to learn from scratch that
\texttt{Nf3} and \texttt{Nd2} share a piece or that \texttt{e4} and \texttt{e5} share a file.
We instead represent every position as a 5-tuple
\[
(\text{kind},\ \text{src},\ \text{dst},\ \text{piece},\ \text{promo}),
\]
where kind $\in\{$PAD, BOS, EOS, MOVE$\}$, src and dst are squares (64 each), piece is one of 6 types and promo one of
5 values.
Tuples are built from \texttt{chess.Move} objects rather than parsed from text, so castling is a king move with real
squares (e1$\to$g1).
The input embedding is the sum of one small table per field.
The output layer has one linear head per field, in the style of AlphaStar's factorized action heads
\citep{vinyals2019alphastar}, and the loss is the sum of the per-field cross-entropies on MOVE positions.

\paragraph{Why factorize.}
A separate vocabulary entry for every (src, dst, piece, promo) tuple would need $64\cdot64\cdot6\cdot5=122{,}880$
entries; at width 960 its input table and output layer alone would hold 236M parameters, more than the whole
decoder.
The four factorized tables and heads hold $(64+64+6+5)\cdot960\cdot2\approx0.27$M.
Legality masks also become cheap per field (squares holding one's own pieces for src, squares not holding them for dst,
piece types still on the board), instead of a mask over every move string.
The cost is expressiveness: the four heads read the same hidden state but are otherwise independent, so the model scores
a move as the sum of four field log-probabilities and cannot directly tie a destination to a particular source square.
Probability that lands on impossible combinations is simply discarded when we keep only the legal moves.

\paragraph{Released tokenizer.}
We release the conversion from any notation to these tokens as a Rust library with Python and WebAssembly bindings,
\hf{chsm8-tokenizer}.
It accepts SAN, LAN, UCI and PGN, mixed within one game, and returns the move tokens or the model
input rows.
Its \texttt{convert(game, src, dst, type)} function also converts between notations: \texttt{src} is \texttt{auto} or
one notation, \texttt{dst} is any notation, numbered PGN, tokens or model rows, and \texttt{type} returns a list or text.

\paragraph{Composing a move.}\label{sec:compose}
At each turn the four heads give log-probabilities over squares, pieces and promotion values.
Each legal move from the board is scored as
\[
\log p(\text{src}) + \log p(\text{dst}) + \log p(\text{piece}) + \log p(\text{promo}),
\]
and a softmax over these scores gives the move distribution; greedy play takes the highest.
A non-promoting move uses promo = none.
A pawn reaching the last rank appears as four legal moves that differ only in the promo field (knight, bishop, rook,
queen), so the promotion head alone decides between queening and underpromotion.

\paragraph{Legality.}
Because only the board's legal moves are scored (python-chess in our evaluation, the Rust library \texttt{shakmaty} in
the demos), every move is legal by construction.
During training we tried two ways of using cheap legality masks (own-piece occupancy and piece types present):
\begin{itemize}\setlength\itemsep{0pt}
\item \textbf{hard mask}: illegal classes are removed from the softmax;
\item \textbf{soft penalty}: logits stay unmasked, and an auxiliary term penalizes probability mass on illegal classes.
\end{itemize}
\Cref{tab:legal} shows that the hard mask backfires against the always-unmasked validation loss: the model never learns
to suppress illegal options itself.
The soft penalty helps slightly up to a weight of 0.5 and hurts beyond it; we use 0.5 throughout pretraining.

\end{verified}
\begin{table}[H]
\centering\small
\begin{tabular}{lr@{\hspace{3em}}lr}
\toprule
\multicolumn{2}{c}{round 1 (30 min, 47.7M model)} & \multicolumn{2}{c}{round 2 (30 min, final architecture)} \\
\cmidrule(r){1-2}\cmidrule(l){3-4}
setting & val loss & soft-penalty weight & val loss \\
\midrule
no legality term & 3.3389 & 0.5 & \textbf{3.3038} \\
hard mask & 6.4273 & 1.0 & 3.3128 \\
soft penalty 0.1 & 3.3379 & 2.0 & 3.3258 \\
soft penalty 0.5 & \textbf{3.3352} & 4.0 & 3.3333 \\
\bottomrule
\end{tabular}
\caption{Illegal-move handling during pretraining (validation loss is always computed without masking; single seed each).
The two rounds use different models, so losses are comparable only within a round.}
\label{tab:legal}
\end{table}

\begin{verified}{pt-2}

\subsection{Architecture}
The model is a Llama-style decoder \citep{touvron2023llama} with RMSNorm, RoPE \citep{su2021roformer}, SwiGLU
\citep{shazeer2020glu} and per-head QK normalization \citep{henry2020qknorm,dehghani2023vit22b}.
Every layer also contains a cross-attention block.
During pretraining it is frozen and fed a constant zero context, so it is a scaffold for later text conditioning rather
than part of the pretrained function.

\paragraph{Search.}
We searched architectures in a fixed wall-clock harness adapted from \texttt{autoresearch}
\citep{karpathy2026autoresearch}, in three rounds with different budgets: 55 minutes per run, then 30 minutes, then a
fixed number of positions.
The first round of 39 runs compared:
\begin{itemize}\setlength\itemsep{0pt}
\item the custom factorized decoder at 3.4M--152M parameters;
\item a from-scratch T5 architecture;
\item pretrained Qwen2.5 (0.5B--7B) and T5 bodies with our tokenization, embedding and output layers, either frozen (only the new
layers trained) or fully fine-tuned.
\end{itemize}
No pretrained language model beat a tuned custom decoder in the same 55 minutes: the best fully fine-tuned Qwen2.5-0.5B
reached a validation loss of 4.13 against 3.67 for a 26M custom decoder, and frozen bodies stayed near 8.
Larger pretrained models also see far fewer positions in a fixed time, so the comparison is about cost as much as
quality.
Learning rate and dropout mattered more than size (\cref{tab:pt-search}).
A second round of 12 thirty-minute runs on 4 GPUs tuned batch size, warmup, kernels and model shape for throughput.
A third, controlled round trained eight $\approx$200M configurations on exactly 862M positions (\cref{tab:arch}).
The 20-layer $\times$768 model had the lowest loss, but the 12-layer $\times$960 model with QK normalization was within
0.011 at 1.8$\times$ the throughput.
We chose it so that a full 78B-position pass fits in about 40 hours on one 4-GPU node (\cref{tab:arch}).

\end{verified}
\begin{table}[H]
\centering\footnotesize\setlength\tabcolsep{4pt}
\begin{tabular}{@{}lrrrrr@{}}
\toprule
configuration (55 min each) & params & trained & positions seen & throughput & val loss \\
\midrule
\multicolumn{6}{@{}l}{\emph{custom factorized decoder}} \\
4L $\times$ 256 & 3.4M & 3.4M & 586M & 178k & 4.959 \\
8L $\times$ 512 & 25.6M & 25.6M & 219M & 66k & 4.417 \\
8L $\times$ 512, lr $6{\times}10^{-4}$ & 25.6M & 25.6M & 244M & 74k & 4.065 \\
\quad + no dropout & 25.6M & 25.6M & 259M & 78k & 4.026 \\
\quad + lr $10^{-3}$ & 25.6M & 25.6M & 409M & 124k & 3.682 \\
\quad + lr $2{\times}10^{-3}$ (best) & 25.6M & 25.6M & 429M & 130k & \textbf{3.668} \\
8L $\times$ 768, lr $10^{-3}$, no dropout & 57.3M & 57.3M & 223M & 68k & 3.688 \\
12L $\times$ 1024, lr $6{\times}10^{-4}$, no dropout & 152.0M & 152.0M & 107M & 32k & 3.846 \\
12L $\times$ 1024, lr $10^{-3}$, no dropout & 152.0M & 152.0M & 108M & 33k & diverged \\
\midrule
\multicolumn{6}{@{}l}{\emph{T5 architecture from scratch}} \\
8L $\times$ 256 & 8.5M & 8.5M & 209M & 63k & 5.239 \\
\midrule
\multicolumn{6}{@{}l}{\emph{pretrained LM body with our tokenization, embedding and output layers}} \\
Qwen2.5-0.5B, fine-tuned, lr $3{\times}10^{-4}$ & 494M & 494M & 29M & 8.8k & 4.134 \\
Qwen2.5-1.5B, fine-tuned & 1.54B & 1.31B & 17M & 5.2k & 4.365 \\
Qwen2.5-0.5B / 1.5B / 3B / 7B, frozen & 0.49--7.07B & 0.3--1.0M & 2--17M & 0.6k--5.2k & 7.90--9.01 \\
T5-small, fine-tuned / frozen & 60.7M & 60.7M / 0.1M & 190--236M & 58k--72k & 6.556 / 7.823 \\
T5-large, frozen & 738M & 0.3M & 38M & 12k & 8.955 \\
\bottomrule
\end{tabular}
\caption{Selected runs from the first pretraining search round (39 runs, 55 minutes each on one node). ``Trained'' counts
the parameters updated; throughput is positions seen per second of the 55-minute budget; frozen bodies train only our tokenization, embedding and output layers. Learning rate and dropout mattered more
than size, and larger models saw far fewer positions in the same time. \Cref{fig:pt-rounds} shows all three search
rounds.}
\label{tab:pt-search}\label{fig:pt-search}
\end{table}

\begin{figure}[H]
\centering
\begin{minipage}[t]{0.49\linewidth}
\centering
\includegraphics[width=\linewidth]{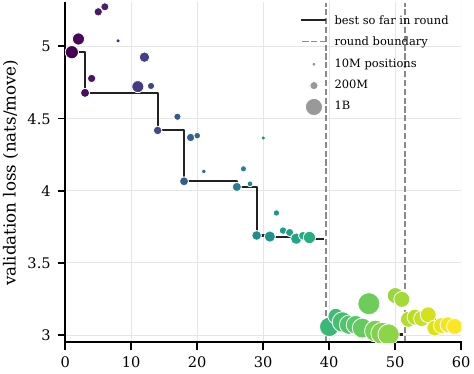}
\caption{Pretraining search in launch order: round 1 (55 min, one node), round 2 (30 min, 4 GPUs) and round 3
(862M positions each). Marker area is the number of positions seen.}
\label{fig:pt-rounds}
\end{minipage}\hfill
\begin{minipage}[t]{0.49\linewidth}
\centering
\includegraphics[width=\linewidth]{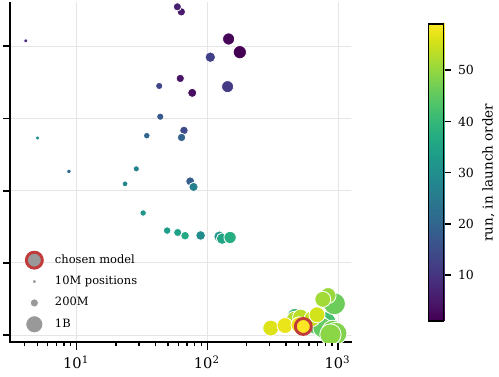}
\caption{Validation loss against throughput (thousand positions per second, log scale) for the same runs; marker area is the
number of positions seen, color the launch order.}
\label{fig:pt-throughput}
\end{minipage}
\end{figure}

\begin{table}[H]
\centering\footnotesize\setlength\tabcolsep{4pt}
\begin{tabular}{@{}lrrrrr@{}}
\toprule
architecture & params (total / trainable) & val loss & positions/s & MFU & full pass \\
\midrule
20L $\times$ 768, SwiGLU 2560, QK-norm & 212.7M / 165.5M & \textbf{3.0485} & 305.5k & 9.8\% & 71\,h \\
\textbf{12L $\times$ 960, SwiGLU 3520, QK-norm} & 210.5M / 166.2M & 3.0598 & 541.8k & 17.1\% & 40\,h \\
20L $\times$ 768, no QK-norm & 212.7M / 165.5M & 3.0660 & 392.3k & 12.5\% & 55\,h \\
12L $\times$ 960, no QK-norm & 210.5M / 166.2M & 3.0710 & 501.4k & 15.8\% & 43\,h \\
20L $\times$ 768, plain FF 3840 & 213.3M / 166.0M & 3.1094 & 463.9k & 14.8\% & 47\,h \\
12L $\times$ 960, plain FF 4800 & 200.1M / 155.8M & 3.1142 & 633.1k & 19.0\% & 34\,h \\
17L $\times$ 768, plain FF 3584 & 174.7M / 134.5M & 3.1244 & 516.7k & 13.5\% & 42\,h \\
10L $\times$ 1024, plain FF 5120 & 189.8M / 147.8M & 3.1407 & \textbf{689.8k} & \textbf{19.6\%} & 32\,h \\
\bottomrule
\end{tabular}
\caption{Fixed-data architecture benchmark: 862M positions each, 4 GH200, identical data and validation batches.
``Total'' includes the frozen cross-attention scaffold. ``Full pass'' is the expected time for our 78.32B-position
pretraining pass on the same node at the measured throughput. The full run of the chosen model held this
throughput for the whole pass: about 530k positions/s (5th--95th percentile 505k--546k across the long job segments),
since the data was packed in advance and I/O stayed constant.}
\label{tab:arch}
\end{table}

\Needspace{16\baselineskip}
\begin{verified}{pt-3}

\subsection{Training run}\label{sec:pt-run}
\paragraph{Recipe.}
\begin{itemize}\setlength\itemsep{0pt}
\item 4 GH200, 32 sequences of 512 positions per GPU, bf16, FlashAttention-3 \citep{shah2024fa3} with variable-length
packing, fused AdamW \citep{loshchilov2019adamw} (weight decay 0.1, clipping 1.0).
\item Peak learning rate $5{\times}10^{-4}$, indexed by positions seen: 5\% linear warmup, then cosine decay to 10\% of
the peak, as in GPT-3 and OLMo~2 \citep{brown2020gpt3,olmo2}.
This replaced an inherited warmup--stable--decay schedule \citep{hu2024minicpm} that had never been ablated here.
\item One pass of 78.32B positions, 1,194,960 optimizer steps, illegal-mass penalty 0.5.
\end{itemize}

\paragraph{Data order.}
The pass used an interleaved curriculum and had to be restarted twice at 60\% because of loss bumps tied to specific
months of the corpus; \cref{sec:curricula} and \cref{fig:pt-loss} describe both.

\end{verified}
\begin{figure}[H]
\centering
\includegraphics[width=\linewidth]{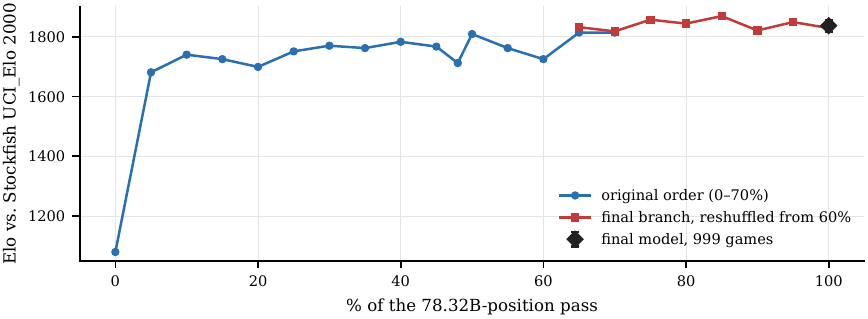}
\caption{Playing strength during pretraining (200 games per checkpoint, $\pm$50 Elo; final point 999 games, 95\% CI). Strength rises quickly to about 1700 within 5\% of the pass and then improves slowly.}
\label{fig:pt-elo}
\end{figure}

\begin{verified}{pt-4}

\subsection{Results}\label{sec:pt-results}
\paragraph{Strength.}
The final model, \hf{chsm8-pt}, scores 139/283/577 (W/D/L) over 999 games, i.e.\ \textbf{1837 Elo} (95\% CI
1817--1856).
\Cref{fig:pt-elo} shows strength during training.

\paragraph{Head to head.}
We played 200 games against each public model, from 100 fixed openings with colors swapped, both sides choosing their
most likely legal move (\cref{tab:versus}).
This comparison is for the base model \pt only.
Like \pt, all opponents are trained only by supervised prediction, without preference tuning, rewards or search, so the
matches compare base models.
The other \chsm models have different goals and are not part of it: \textsc{chsm8-player-base} and the player
fine-tunes are also trained only on moves, with no reward, but the player models aim to imitate one player rather than
to play as well as possible; \sft and \dpo are trained to follow a style description, not to find the strongest move.
Maia-2 predicts human moves from the board; Chess-GPT, the ChessFormer from the Transcendence study and chess-gpt2-uci are next-token models on
game records (the last one does not document its data); ChessGPT-base is a language model pretrained on chess text,
including games, books and forum posts.
\chsm wins every match, including against ChessGPT-base, which has 13 times as many parameters; model size alone does
not explain the ranking.
The Transcendence ChessFormer was trained only on players rated $\le1500$ to study transcendence \citep{zhang2024transcendence}, so a
large gap is expected there.
\Cref{sec:versus1k} extends these matches to 1,000 games and to recent strong models, including Maia-3, Allie,
DeepMind's searchless transformers and Matilda, which uses Stockfish.

\end{verified}
\begin{table}[H]
\centering\small
\resizebox{\linewidth}{!}{\begin{tabular}{llrrrr}
\toprule
opponent & type & parameters & W / D / L & score & Elo advantage (95\% CI) \\
\midrule
Maia-2 blitz, rating input 2000 \citep{tang2024maia2} & board policy & 23M & 102 / 52 / 46 & 0.640 & +100 (+59, +144) \\
Chess-GPT, 16 layers \citep{karvonen2024chessgpt} & char-level PGN & 51M & 123 / 43 / 34 & 0.723 & +166 (+122, +216) \\
chess-gpt2-uci-12x12x768 & UCI-text GPT-2 & 86M & 151 / 37 / 12 & 0.848 & +298 (+249, +359) \\
ChessGPT-base-v1 \citep{feng2023chessgpt} & LM on chess text & 2.8B & 158 / 32 / 10 & 0.870 & +330 (+278, +397) \\
ChessFormer (Transcendence) \citep{zhang2024transcendence}, $\le$1500 data & char-level PGN & 51M & 163 / 30 / 7 & 0.890 & +363 (+309, +434) \\
\bottomrule
\end{tabular}}
\caption{\pt head to head, 200 games per opponent, greedy legal moves on both sides, no search.}
\label{tab:versus}
\end{table}

\begin{figure}[H]
\centering
\includegraphics[width=\linewidth]{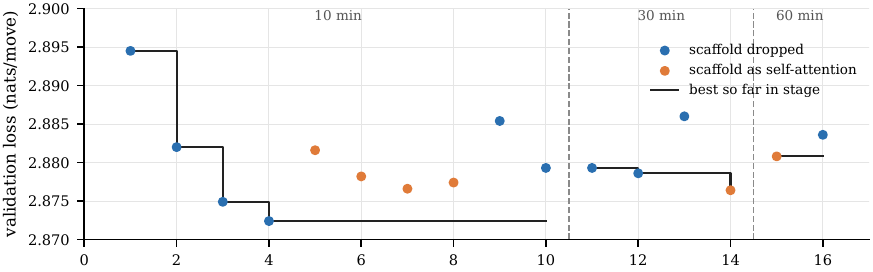}
\caption{Games-only continuation search in launch order, in stages of 10, 30 and 60 minutes. Blue: scaffold dropped
(its constant output becomes a per-layer bias). Orange: a gated second self-attention in its place, copied or random;
slightly lower loss at 30 minutes but 35--40\% slower and no better at 60, so the full pass drops the scaffold.}
\label{fig:games-search}
\end{figure}

\begin{verified}{pt-5}

\paragraph{Games-only continuation and player models.}
For player-specific models we continued \pt without the cross-attention scaffold for one pass over 26.0M unique games;
the result is \hf{chsm8-player-base} (\textbf{1890 Elo}).
A 16-run search (\cref{fig:games-search}) chose to drop the cross-attention scaffold and train at learning rate
$5{\times}10^{-5}$; longer search runs did not lower validation loss further.

\end{verified}

\begin{verified}{pt-6}

From it we fine-tuned 14 player models (10 current top players and 4 historical ones) on each player's own moves.
After comparing full fine-tunes, LoRA \citep{hu2022lora} and DPO on Nakamura's games, we chose a full fine-tune at
learning rate $3{\times}10^{-5}$ for 2 passes.
A player has only $10^3$--$10^4$ games, so a fine-tune must repeat them; the number of passes was set on Nakamura's
held-out games (base: 54.1\% test move match, 1886 Elo).
We kept the recipe with the best validation move match among those within 75 Elo of the base:
two passes at $3{\times}10^{-5}$ reached 57.2\% (1846 Elo).
Four passes at $10^{-5}$ were no better (57.2\%, 1856), two passes at $3{\times}10^{-6}$ fitted less (55.6\%),
LoRA rank 16 for three passes reached 56.8\% (1842), and more LoRA passes or higher rates diverged in cross-entropy and
lost strength (rank 64, six passes: 1741 Elo).
Move-match accuracy on held-out games changes by $-0.5$ to $+3.2$ points (up for 13 of 14 players), and strength stays within $\pm$40 Elo of the base.
Each player model played 1,000 games against Stockfish (95\% CI about $\pm$19 Elo); all but two are within the
base model's interval (1870--1909): Wesley So is 1 Elo above it (1910) and Nakamura below it, so these Elo values
say only that no strength was lost, except for Nakamura.
Nakamura is the instructive exception: his fine-tune moved furthest toward its player (held-out cross-entropy 3.039 to
2.862, the largest drop of the 14) and lost the most strength (1850 Elo against 1890 for the base).
Fitting a specific style more closely costs strength against an engine, as with style prompts in the main pipeline
(\cref{sec:results}).
DPO-based player fits were unstable: SFT followed by DPO collapsed to 1226--1374 Elo.
\end{verified}

\section{Style labels}\label{sec:labeler}
\begin{verified}{labels-1}
\subsection{Task and prompt}
The labeler sees one side of one game.
Its input is the SAN move list plus a \emph{feature card} measured from the board: development speed, castling side and
timing, captures, checks, sacrifices, material deficit, central control, pawn moves and pawn storms, king pressure,
mobility, trades, queen trade and endgame share.
Player names and the result are hidden.
The output is five descriptions of increasing length, written as instructions a user might type: two terse, one short,
one medium and one long.
\Cref{app:examples} shows an input card and labels.

\paragraph{The feature card is essential.}
We scored prompts with automatic metrics:
\begin{itemize}\setlength\itemsep{0pt}
\item \textbf{grounding}: can an LLM judge pick the described game out of four?
\item \textbf{distinctiveness}: are labels of different games different?
\item \textbf{fidelity}: do claims in the label correlate with measured features?
\item \textbf{leakage}: does the label quote moves, squares or numbers?
\end{itemize}
Without the card, grounding fell to 0.29 against a 0.25 chance level, and labels were often false.
With the card but no moves, labels were true but interchangeable.
Both together worked best (grounding 0.74).
An evolutionary prompt search over three rounds produced prompts that won on the games they were optimized on but lost
on fresh games, so we kept a hand-written prompt (\texttt{s10-mini}).

\subsection{Teacher models}\label{sec:teachers}
All candidate teachers were queried through the SwissAI serving platform at CSCS
(\url{https://serving.swissai.svc.cscs.ch/}), which hosts open models for free; labeling millions of games on a paid
API was never an option.
We screened them with the prompt, the metrics above and manual reading of the outputs on the same held-out games
(\cref{tab:teachers}):
\begin{itemize}\setlength\itemsep{0pt}
\item \textbf{Gemma-4-31B}: grounded, varied and parse-clean (fitness 0.58, 97\% clean, 100\% parsed);
kept.
\item \textbf{Apertus-v1.5-70B}: a not yet released checkpoint of the Apertus family \citep{apertus2025}
(\url{https://huggingface.co/swiss-ai}) served on the same platform.
Its automatic scores were mediocre, but read by eye its labels were the most vivid of all candidates, and serving it
on its own nodes alongside Gemma added labeling capacity.
We therefore kept it as a second teacher.
The released Apertus-70B-Instruct-2509 (v1) and Apertus-v1.5-8B were screened as well: v1 wrote more distinctive labels
but made the most false claims, and the 8B model was weaker.
\item \textbf{MiniMax-M2.5}: the most distinctive labels in the first screen, but it ignores the switch that turns
thinking off and spends its token budget on visible reasoning; with a large enough budget it is slow, and its final
labels invent events (sacrifices, rook endings) that the card rules out.
\item \textbf{Qwen2.5-72B}: parses, but falls back to the generic ``develop, control the center, attack the kingside''
template and invents sacrifices under every sampling setting.
\item \textbf{Llama-3.3-70B}: low distinctiveness (0.20 against 0.39 for Gemma).
\item \textbf{GLM-4.7-Flash}: best retrieval in one screen, but parsed only 25--75\% of outputs at 200\,s latency.
\item \textbf{Ministral-3-14B}: parsed one output in three, with broken mixed-language text, and its endpoint rejected
sampling controls.
\item \textbf{Paid frontier models} (GPT-5.2, GPT-5-mini, GPT-5-nano, Grok-4.5): none beat Gemma-4-31B; GPT-5-nano
wrote nearly identical labels for unrelated games (distinctiveness 0.07).
\end{itemize}
\end{verified}
\begin{table}[H]
\centering\footnotesize\setlength\tabcolsep{4pt}
\begin{tabular}{@{}lrrrr@{}}
\toprule
teacher & fitness & distinctiveness & retrieval R@1 & parsed \\
\midrule
MiniMax-M2.5 & \textbf{0.598} & \textbf{0.515} & 0.67 & 100\% \\
\textbf{Gemma-4-31B} & 0.579 & 0.387 & 0.53 & 100\% \\
Grok-4.5 & 0.568 & 0.335 & 0.48 & 100\% \\
GPT-5.2 & 0.546 & 0.301 & 0.45 & 100\% \\
GPT-5-mini & 0.533 & 0.311 & 0.46 & 92\% \\
Apertus-70B-Instruct-2509 (v1) & 0.441 & 0.357 & --- & 100\% \\
Llama-3.3-70B & 0.429 & 0.198 & 0.35 & 100\% \\
GLM-4.7-Flash & 0.404 & 0.501 & \textbf{0.73} & 25\% \\
GPT-5-nano & 0.339 & 0.066 & 0.20 & 67\% \\
\textbf{Apertus-v1.5-70B} & 0.332 & 0.133 & --- & 100\% \\
Apertus-v1.5-8B & 0.306 & 0.163 & --- & 100\% \\
Ministral-3-14B & 0.240 & 0.237 & 0.39 & 33\% \\
\bottomrule
\end{tabular}
\caption{Teacher screening with the same prompt on the same held-out games. Fitness combines distinctiveness,
fidelity and grounding; R@1 is how often a judge retrieves the described game. The Apertus models were screened in a
later round, without the retrieval test. Bold names are the two production teachers. The scores rank the models,
but read side by side the labels of the stronger models looked very similar, so the final choice was made by reading
them.}
\label{tab:teachers}
\end{table}

\begin{verified}{labels-2}

In practice the automatic scores separated only the clearly broken models; the labels of the remaining candidates
looked very similar to each other, and we chose the teachers by reading their labels side by side.
Gemma-4-31B and Apertus-v1.5-70B were the two that wrote the kind of description we wanted: vivid and specific, close to what
a person would type to describe how a game was played.
Production labeling therefore used both: Gemma-4-31B for 3.02M sides and Apertus-v1.5-70B, with thinking on and off, for
4.75M.
On the platform each teacher ran on several nodes at once, so its total rate depended on how many were serving it.
On a single 4-GPU node of our own in FP8, Gemma-4-31B labeled 7.5 sets/s on average and 12.5 at best; this per-node
rate is the one used for the cost estimates below.

\paragraph{Gemma is the better training signal.}
Read by eye, Apertus labels are as good as Gemma's; measured against the card they are not.
A label is \emph{contradicted} when it makes a claim that the feature card rules out, such as a sacrifice when there
was none, a king attack with no king pressure, a queen trade when the queens stayed on, slow development when it was
fast, or ``never castle'' when the side castled (the regex check in \cref{sec:distill}).
We report this at two levels.
The \emph{label} rate is the share of individual descriptions (bands) that are contradicted.
The \emph{set} rate is the share of five-band sets with at least one contradicted band; it is always higher, because
one bad band out of five is enough.
For example, 4.5\% of Gemma's labels are contradicted, but about 20\% of its sets contain at least one such label, and
after relabeling with the clean student the rates are 0.40\% of labels and 1.9\% of sets.
\begin{itemize}\setlength\itemsep{0pt}
\item 18.2\% of Apertus labels contradict the card on the audit games, against 4.6--4.7\% for Gemma; most often
Apertus describes a sacrifice that never happened (in 58 of 76 audited games).
\item Across the whole pool, 16.5\% of Apertus bands are contradicted, against 4.6\% for Gemma.
\item Students trained on the mixed pool copied this: 14--19\% of their labels were contradicted, against about 4--5\%
when trained on Gemma alone.
\item SFT models trained on each source's labels show that Apertus labels carry 4--5$\times$ less information about the
game (\cref{sec:label-signal}).
\end{itemize}
We therefore distilled the labeler from Gemma-4-31B labels only.

\subsection{Distilled labeler (\hf{chsm8-labeler-gemma3-1b})}\label{sec:distill}
We first labeled with Gemma-4-31B itself.
At about 7.5 sides per second per 4-GPU node, the remaining 22.3M sides would have needed about 830 node-hours, more
than a month on one node and days even on several.
Instead we distilled the teacher into a 1B student, which then labeled all 22.3M sides in about 4.5 hours on three
nodes (12 GH200), about 14 node-hours.
Students were fine-tuned for 65 minutes on one GH200 each and judged on 384 held-out test games
(\cref{tab:labeler}).
We used three quality measures:
\begin{itemize}\setlength\itemsep{0pt}
\item a \textbf{regex check} that flags five claim types contradicted by the card: a sacrifice when there was none, a
king attack with no king pressure, a queen trade when the queens stayed on, slow development when it was fast, and ``never
castle'' when the side castled;
\item \textbf{claim grounding}: Qwen3-32B extracts claims on 12 style axes, and we measure the share of checkable claims
that the card supports;
\item \textbf{distillation}: Cohen's $\kappa$ \citep{cohen1960kappa} between student and teacher claims on the same game,
minus that on a different game, relative to the same quantity for a second teacher sample.
\end{itemize}

\end{verified}
\begin{table}[H]
\centering\small
\resizebox{\linewidth}{!}{\begin{tabular}{lrrrr}
\toprule
model & labels contradicted & claim grounding & distillation & vLLM games/s \\
\midrule
Gemma-4-31B teacher & 4.7\% & 88.3\% & --- & --- \\
Gemma-4-31B, second sample (ceiling) & --- & 87.9\% & 100\% & --- \\
\textbf{Gemma-3-1B, full fine-tune} & 4.7\% & 88.2\% & \textbf{94\%} & \textbf{154} \\
Gemma-3-270M, full fine-tune & 5.4\% & 88.3\% & 90\% & 193 \\
T5Gemma 2B-2B, LoRA & 5.5\% & 87.4\% & 94\% & 23 (HF) \\
Qwen3-8B, LoRA & 4.0\% & 87.9\% & 92\% & 40 \\
Gemma-4-E4B, LoRA & 4.0\% & \textbf{89.2\%} & 87\% & 59 \\
T5Gemma 9B-2B, LoRA & 4.3\% & 87.3\% & 86\% & --- \\
\bottomrule
\end{tabular}}
\caption{Labeler students after 65 minutes of fine-tuning on Gemma-4-31B labels (384 test games, single GH200).}
\label{tab:labeler}
\end{table}

\begin{verified}{labels-3}

All students reached teacher-level grounding, so speed decided: a full fine-tune of Gemma-3-1B \citep{gemma3} ties for the
best distillation score and labels 154 games/s with vLLM.
An earlier apparent advantage of encoder--decoder students came from training on mixed labels: decoder students copied
Apertus's invented sacrifices more.
Two families were dropped in an earlier screen with a different protocol (128 validation games, unoptimized
generation), so they are not in \cref{tab:labeler}.
LLaDA-8B \citep{nie2025llada} was well grounded (3.4--3.9\% of labels contradicted, against 3.8\% for Gemma-3-1B in
the same screen) but parsed only 89--91\% of outputs and generated 0.14 games/s, 25$\times$ slower than T5Gemma 2B-2B.
FLAN-T5-XXL and FLAN-UL2 \citep{chung2022flan} reached a validation loss of 0.41 and 0.39 against 0.34 for
T5Gemma 2B-2B under the same training.

\paragraph{Full training and repair.}
Trained for one epoch on all 3.02M Gemma sets, the student's distillation score reached 106\%, above the teacher's
self-agreement, because greedy decoding gives the teacher's most likely claims without sampling noise.
It also copied the teacher's error rate (4.5\% contradicted labels).
We therefore retrained it only on the 80\% of Gemma sets with no contradicted band: 2,402,472 training sets, with
1,524 validation and 1,517 test sets left unfiltered, released as \hfd{chsm8-distillation}.
This ``clean'' model contradicts the card in 0.5\% of labels, against 4.7\% for the teacher.
All 30.1M sets were then checked; a set with one bad band had that band regenerated given the game and the earlier bands,
and a set with more had the whole set regenerated.
After five repair rounds, 4,868 of about 150M bands (0.003\%) remain flagged.

\end{verified}

\begin{verified}{labels-4}

\subsection{Label quality matters more than volume}\label{sec:label-signal}
We trained SFT models on 1.75M games of \emph{pure} labels from each source (no repaired band, no flagged band) and
tested on pure labels from each source (\cref{tab:labelsrc}).
The right-minus-wrong cross-entropy gap measures how much a description tells the model about the game.
Crossing training and test sources answers two separate questions.
The diagonal asks how much signal a source carries for a model trained on it; but a model can also learn to read its
own labeler's habits, so the off-diagonal cells ask whether what it learned transfers to descriptions written by
another labeler.
This matters because the final model is trained on a mix of teacher and student labels and is then prompted by users,
whose wording comes from neither.

\end{verified}
\begin{table}[H]
\centering\small
\begin{tabular}{lrrr}
\toprule
trained on $\downarrow$ / tested on $\rightarrow$ & Gemma-4-31B & Apertus-v1.5-70B & distilled Gemma-3-1B \\
\midrule
Gemma-4-31B & $-0.061$ & $-0.015$ & $-0.068$ \\
Apertus-v1.5-70B & $-0.014$ & $-0.013$ & $-0.015$ \\
distilled Gemma-3-1B & $-0.062$ & $-0.017$ & $\mathbf{-0.072}$ \\
\bottomrule
\end{tabular}
\caption{Right-minus-wrong description cross-entropy (nats per move; more negative means the text is used more), 1.75M
training games per source.}
\label{tab:labelsrc}
\end{table}

\begin{verified}{labels-5}

Apertus labels carry 4--5$\times$ less signal whichever way they are used.
It describes nearly every game as ``attack the king, push pawns, trade, sacrifice'' in fresh wording: ``sacrifice''
appears in 19\% of medium labels against 6\% for Gemma.
The distilled student is as informative as its teacher.
All 4.75M Apertus sets were relabeled with the clean student, and the final SFT model was trained on the relabeled data.
On these relabeled sides 0.40\% of bands contradict the card (1.9\% of sets have at least one), against 16.5\% for
Apertus's own labels of the same sides.
The low rate comes from training the student only on pure labels: the clean model saw only the 2.4M Gemma sets in which
no band is contradicted, while the earlier model trained on all Gemma sets copied the teacher's 4.5\%.
The public \hfd{chsm8-post-training} release contains exactly these relabeled sets, with the splits used
for SFT and DPO.
\end{verified}

\paragraph{Filtering Apertus did not help.}
Before distilling from Gemma alone, we tried a \emph{clean mix}: all Gemma sets plus the 42\% of Apertus sets that pass
the regex check.
A student trained on it reached a worse validation loss than one trained on Gemma alone (0.269 against 0.143 nats per
byte), with no measurable gain in grounding.
The regex check covers only five claim types, so Apertus's other invented events pass it.

\begin{figure}[H]
\centering
\includegraphics[width=\linewidth]{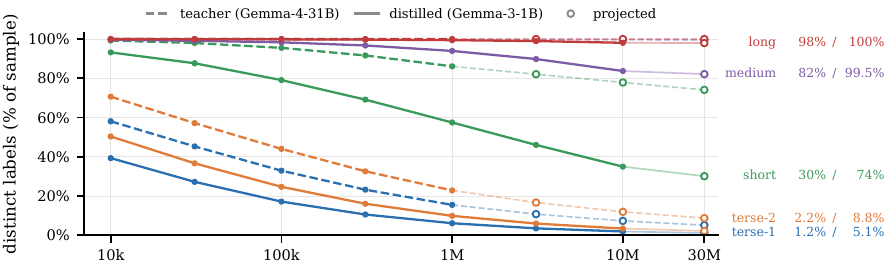}
\caption{Distinct label strings per band against the number of label sets sampled. Dashed: Gemma-4-31B teacher (up to
1M of its 2.4M sets); solid: distilled Gemma-3-1B; circles: power-law projection, an upper bound. Labels give the
projected share at 30M sets, distilled / teacher.}
\label{fig:label-uniqueness}
\end{figure}

\paragraph{Label diversity.}
The distilled labeler writes less varied text than its teacher (\cref{fig:label-uniqueness}).
In samples of 1M label sets, 6--10\% of the student's terse tags are distinct, against 16--23\% for Gemma-4-31B, and
58\% of its short labels, against 86\%.
Projected to all 30M sets, the terse bands behave like a vocabulary of a few hundred thousand phrases (360k--670k
distinct), while medium and long descriptions stay nearly unique (82\% and 98\%).
This does not cost signal: the student's labels tell the model as much about a game as the teacher's
(\cref{tab:labelsrc}), because different games still get descriptions that fit them.

\section{Style SFT}\label{sec:sft}
\begin{verified}{sft-1}
All SFT models start from \pt and share three design choices:
\begin{itemize}\setlength\itemsep{0pt}
\item \textbf{frozen text encoder}: the description is embedded by a frozen pretrained encoder, so the text model stays a
separate, replaceable module;
\item \textbf{gated cross-attention}: the scaffold's cross-attention blocks in the top layers attend to the text, and their
output is added as $x+\tanh(g)\cdot\mathrm{attn}(x,c)$ with $g$ initialized at 0 \citep{alayrac2022flamingo}, so
training starts exactly at the pretrained model;
\item \textbf{described-side loss}: the loss is the pretraining objective restricted to the moves of the described side;
opponent moves are context only.
\end{itemize}
Layers below the first conditioned layer never see the text.
They run once per game and are shared across all descriptions of that game.
\Cref{fig:arch} shows the resulting model.
\end{verified}
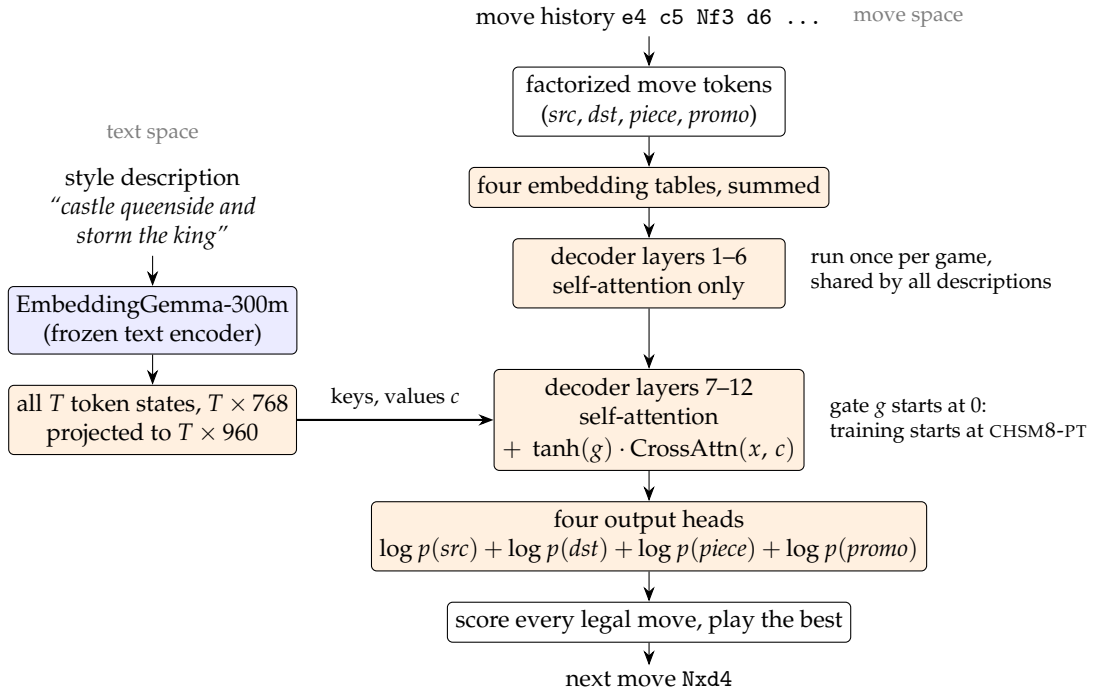
\begin{figure}[H]
\centering
\begin{tikzpicture}[
  font=\footnotesize, >={Stealth[length=2mm]}, node distance=4mm,
  box/.style={draw, rounded corners=2pt, align=center, inner sep=3pt, minimum width=36mm},
  frozen/.style={box, fill=blue!8},
  train/.style={box, fill=orange!12},
  io/.style={align=center, inner sep=2pt}]
\node[io] (hist) {move history \texttt{e4 c5 Nf3 d6 \ldots}};
\node[box, below=of hist] (tok) {factorized move tokens\\(\textit{src}, \textit{dst}, \textit{piece}, \textit{promo})};
\node[train, below=of tok] (emb) {four embedding tables, summed};
\node[train, below=of emb] (low) {decoder layers 1--6\\self-attention only};
\node[train, below=8mm of low, minimum height=12mm] (high) {decoder layers 7--12\\self-attention\\[1pt]
  $+\;\tanh(g)\cdot\mathrm{CrossAttn}(x,\,c)$};
\node[train, below=of high] (heads) {four output heads\\$\log p(\textit{src})+\log p(\textit{dst})+\log p(\textit{piece})+\log p(\textit{promo})$};
\node[box, below=of heads] (legal) {score every legal move, play the best};
\node[io, below=3mm of legal] (move) {next move \texttt{Nxd4}};
\node[train, left=26mm of high] (proj) {all $T$ token states, $T\times768$\\projected to $T\times960$};
\node[frozen, above=of proj] (enc) {EmbeddingGemma-300m\\(frozen text encoder)};
\node[io, above=of enc] (text) {style description\\\emph{``castle queenside and}\\\emph{storm the king''}};
\draw[->] (hist) -- (tok); \draw[->] (tok) -- (emb); \draw[->] (emb) -- (low); \draw[->] (low) -- (high);
\draw[->] (high) -- (heads); \draw[->] (heads) -- (legal); \draw[->] (legal) -- (move);
\draw[->] (text) -- (enc); \draw[->] (enc) -- (proj);
\draw[->, thick] (proj.east) -- node[above, font=\scriptsize] {keys, values $c$} (high.west);
\node[font=\scriptsize, gray, above=1mm of text] {text space};
\node[font=\scriptsize, gray, right=2mm of hist] {move space};
\node[font=\scriptsize, align=left, right=2mm of low] {run once per game,\\shared by all descriptions};
\node[font=\scriptsize, align=left, right=2mm of high] {gate $g$ starts at 0:\\training starts at \pt};
\end{tikzpicture}
\caption{The style model: the top six decoder layers attend to the frozen encoder's token states through gated
cross-attention. Blue: frozen; orange: trained in SFT.}
\label{fig:arch}
\end{figure}
\begin{verified}{sft-2}
\subsection{Final model (\hf{chsm8-sft})}\label{sec:sft-final}
\paragraph{Recipe.}
\begin{itemize}\setlength\itemsep{0pt}
\item frozen EmbeddingGemma-300m \citep{vera2025embeddinggemma} with all $T\times768$ token states as keys, projected to
the decoder width of 960 and encoded online;
\item conditioning on the top 6 of 12 decoder layers, all 12 layers trained;
\item the five LLM bands (no metadata labels), Gemma-4-31B plus distilled labels, with all Apertus sets relabeled;
\item 10\% label dropout and a pinned no-text term;
\item top-player games at a flat 6\% of batches over three passes of the player corpus, with fresh wording each pass;
\item learning rates: decoder $5{\times}10^{-7}$, text cross-attention modules (the text projection and the gated
cross-attention blocks) $8{\times}10^{-4}$, cosine decay to 0 after 5\% warmup; 128 games per GPU.
\end{itemize}
Training ran for one pass: 28,239 steps on 8 GH200 at about 2,400 games/s, 3.1 hours.
\Cref{sec:sft-search,sec:sft-diag} describe how each choice was made.

\paragraph{Result.}
Test cross-entropy is 2.811 with the right description, 3.115 with a wrong one and 2.946 with none.
The right--wrong gap is $-0.303$, six times the $-0.049$ of the first full run with pooled conditioning (\cref{sec:sft-diag}).
Halfway through training, per-token SFT already followed style prompts in play as well as the first full run \emph{after}
DPO: 98\% of castling went queenside under \pc, against 39\% for the first full run and 74\% for it after DPO.

\paragraph{Player prompts.}
We scored the player's own moves in held-out games of the top players under a \emph{``play like \{player\}''} prompt
(one random template and alias per game), under the same wording with another top player's name, and with no prompt
(\cref{tab:playerprompts}).
We used all 99 players, 30 held-out games each (2,768 games, 124k moves).
Move match is how often the model's top legal move is the move the player actually played.
A uniformly random legal move would match 5.8\% of the time; the model already matches 54.7\% with no prompt, because
strong players mostly play the moves any strong model expects.
The prompt acts on the remaining moves: with the right name, SFT matches 56.2\%, against 55.0\% with another player's
name, and cross-entropy falls from 2.962 to 2.850.
The right name beats the wrong one in cross-entropy for 94 of 99 players (DPO: 96) and in move match for 88 (DPO: 88).
\Cref{app:players} compares the prompted models with the dedicated player fine-tunes.
DPO follows player names slightly less well than SFT (move match 55.5\% vs 56.2\% on all players): its preference
pairs come from the style descriptions only, and a wrong name hurts it more (cross-entropy 3.047 vs 2.962).
\end{verified}
\begin{table}[H]
\centering\footnotesize\setlength\tabcolsep{4pt}
\begin{tabular}{@{}lrr@{}}
\toprule
model, prompt & CE & move match \\
\midrule
\sft, right name & \textbf{2.850} & \textbf{56.2\%} \\
\sft, another player's name & 2.962 & 55.0\% \\
\sft, no prompt & 2.984 & 54.7\% \\
\dpo, right name & 2.879 & 55.5\% \\
\dpo, another player's name & 3.047 & 53.9\% \\
\dpo, no prompt & 2.984 & 54.6\% \\
\textsc{chsm8-player-base} (no text) & 2.919 & 55.8\% \\
uniformly random legal move & --- & 5.8\% \\
\bottomrule
\end{tabular}
\caption{Player prompts on held-out games of all 99 top players (2,768 games): cross-entropy (nats per move) and top-1
legal move match on the player's own moves. Per-game standard deviation of move match is about 9 points, so the
standard error is about 0.2 points.}
\label{tab:playerprompts}
\end{table}

\begin{figure}[H]
\centering
\includegraphics[width=\linewidth]{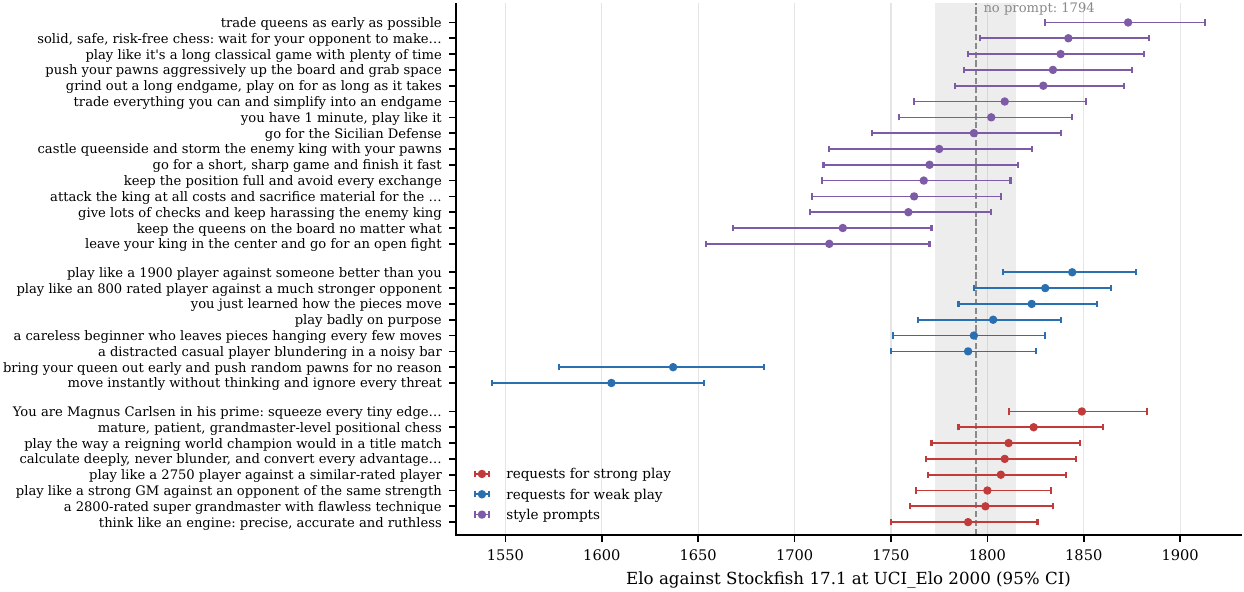}
\caption{Final SFT model against Stockfish under each evaluation prompt (200 games per style prompt, 300 per request
for strong or weak play). Gray band: 95\% interval with no prompt (1,000 games).}
\label{fig:sft-prompt-elo}
\end{figure}

\paragraph{Strength under prompts.}
Asking the final SFT model for strong or weak play does not change its strength (\cref{fig:sft-prompt-elo}): even
\emph{``play like an 800 rated player''} and \emph{``play badly on purpose''} stay within the no-prompt interval, because no
training description was ever paired with weak play.
Only prompts that name concrete bad moves lower it: \emph{``move instantly without thinking and ignore every threat''}
plays at 1605 and \emph{``bring your queen out early and push random pawns''} at 1637, against 1794 with no prompt
(1,000 games; 200--300 per prompt).

\begin{figure}[H]
\centering
\includegraphics[width=\linewidth]{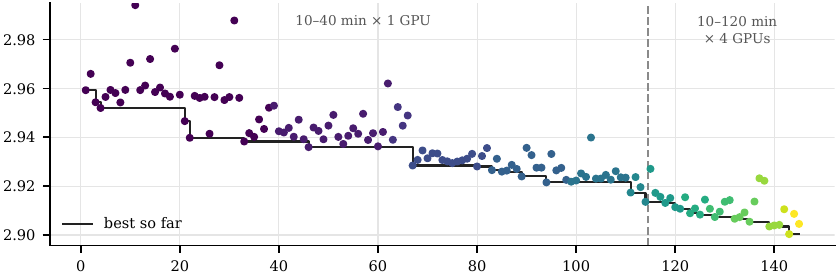}
\caption{SFT recipe search: validation cross-entropy on the LLM descriptions (the selection metric) for every run,
on the same 4,000 validation games, in launch order: first 10--40-minute runs on one GPU, then 10--120-minute runs on
four GPUs (right of the dashed line). Each round was planned from the previous ones; the line is the best so far.}
\label{fig:sft-search}
\end{figure}

\begin{verified}{sft-3}
\paragraph{The alignment tax.}
With no prompt, the final SFT model plays at 1792 Elo (2,000 games), about 45 below \pt (1837) on the same protocol;
the first full run was at 1828.
SFT tunes the decoder toward the style descriptions, but on games of strong players only, so it stays strong; still,
part of its capacity moves from predicting the best move to following the description, and plain move prediction gets
slightly worse.
The pinned no-text term freezes only the empty-prompt cross-attention output, not the decoder weights beneath it, so it
does not prevent this.
The tax is larger for the final model, whose conditioning is much stronger than in the first full run.

\subsection{Recipe search}\label{sec:sft-search}
About 145 fixed-budget runs on the first 7.8M labeled sides, in 14 rounds, set most of the recipe
(\cref{fig:sft-search}).
The early rounds used \emph{pooled} conditioning: each description is compressed into a single text vector, which
is the only key the cross-attention can attend to.
\begin{itemize}\setlength\itemsep{0pt}
\item \textbf{encoders}: the final model uses EmbeddingGemma-300m with per-token keys (\cref{sec:sft-diag}).
Among pooled encoders, Qwen3-Embedding-0.6B \citep{zhang2025qwen3emb}, last-token pooled, was best; at equal games per
step and a 10-minute budget it reached a description cross-entropy of 2.9385, against 2.9432 for MiniLM
\citep{wang2020minilm}, 2.9438 for BGE \citep{xiao2023bge} and 2.9425 for per-token states of the Qwen3-0.6B language
model.
T5Gemma encoders were worse (2.9500 and 2.9617 for 2B and 9B) and 2.5--6$\times$ slower, and Qwen3-Embedding-4B gave
no gain over 0.6B (2.9235 against 2.9226--2.9231).
\item \textbf{depth}: condition the top 6 of 12 layers and train all 12; the top 6 beat all 12 and was faster.
\item \textbf{rejected conditioning}: FiLM \citep{perez2018film} was worse, and a frozen decoder was much worse.
\item \textbf{batches}: 128 games per GPU, length-bucketed (0.9\% padding); data held in RAM, because random Lustre reads
cost 43\,ms per game.
\end{itemize}

\paragraph{Short-horizon bias.}
Learning rates tuned in 10-minute runs got worse at 30--120 minutes.
The best decoder learning rate fell from $2{\times}10^{-6}$ at 10 minutes to $10^{-6}$ at 60 minutes and
$5{\times}10^{-7}$ at 120 minutes, while the cross-attention rate could stay high.
We therefore tuned at the target length.
\end{verified}

\begin{figure}[H]
\centering
\includegraphics[width=\linewidth]{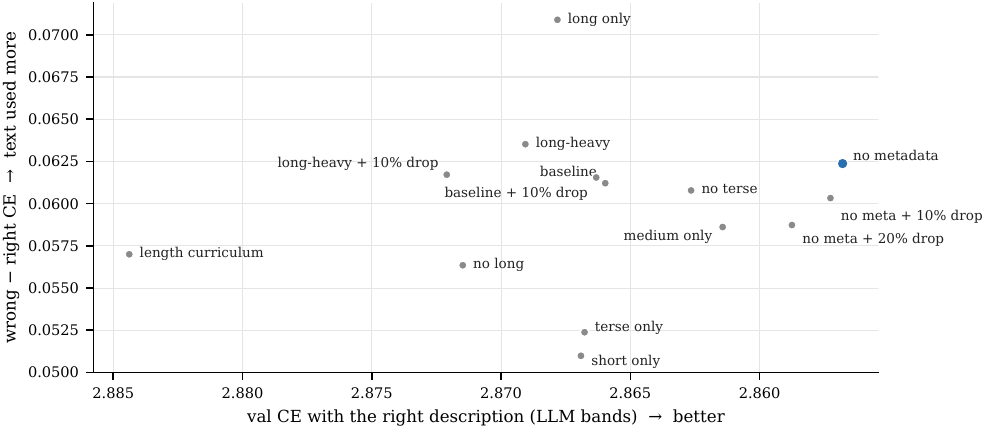}
\caption{Label-band variants, 1 hour each on the same 4M games, compared on their last periodic validation (steps
6,000--7,000). Right is better (lower cross-entropy with the right description), up is better (larger wrong minus right
cross-entropy, so the text is used more); blue is the recipe we kept. The variants drop one band (\emph{no terse},
\emph{no long}), train on a single band (\emph{terse only}, \emph{short only}, \emph{medium only}, \emph{long only}),
sample long bands more often (\emph{long-heavy}), drop the four metadata labels (\emph{no metadata}), order training
from short to long bands (\emph{length curriculum}), or add 10--20\% label dropout (\emph{+ drop}).}
\label{fig:sft-bands}
\end{figure}
\begin{figure}[H]
\centering
\begin{minipage}[c]{0.6\linewidth}
\includegraphics[width=\linewidth]{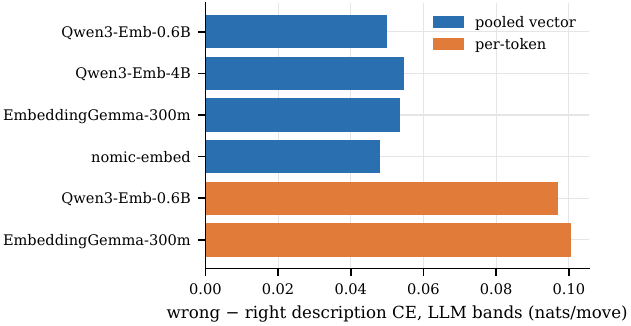}
\end{minipage}\hfill
\begin{minipage}[c]{0.36\linewidth}
\caption{Wrong minus right cross-entropy on the LLM bands for pooled and per-token conditioning (same 500k games and
schedule).}
\label{fig:sft-enc}
\end{minipage}
\end{figure}

\begin{verified}{sft-4}
\paragraph{Other findings.}
\begin{itemize}\setlength\itemsep{1pt}
\item \textbf{Labels}: Apertus labels carried little signal (\cref{sec:label-signal}).
\item \textbf{Metadata labels} overfitted their template wording; dropping them lowered LLM-label cross-entropy by 0.009.
\item \textbf{Bands}: all five bands help (\cref{fig:sft-bands}).
The long band carries the most signal: dropping it raised cross-entropy (2.872 against 2.857 with all
bands) and a smaller gap (0.056 against 0.062), while training on it alone gave the largest gap (0.071) but the worst
cross-entropy on terse prompts (2.872 against 2.851).
Any single short band alone gave the smallest gaps (0.051--0.052).
Reweighting toward long bands bought gap at a cost in cross-entropy, and the length curriculum was worst on both
(2.884, 0.057).
\item \textbf{Label dropout} of 10\% improved the no-text path at almost no cost (cross-entropy $+0.0005$, gap $-0.002$);
20\% cost more of the gap ($-0.004$).
\item \textbf{Players and repeated passes}: what mattered was the share of player games, not the number of wordings
(75-minute runs on a 4M-side subset).
A flat 10\% share gave a 5$\times$ larger right-vs.-wrong player gap than a schedule that read every player game once,
at about 1\% until 80\% of the run and rising to 9.5\% at the end.
Raising the share to 20\% cost the description prompts (validation cross-entropy $+0.005$, smaller right-vs.-wrong gap)
for a small player gain.
One, two or four distinct wordings per player game were within noise ($\pm0.003$).
On the full corpus a single pass over the 542k player records is only 2\% of the 27.08M sides, so a flat share of
about 6\% needs three passes; since wording count did not matter, each pass simply draws a fresh template and alias.
\item \textbf{A drifting no-text path}: the empty-context cross-attention output ($\mathrm{out\_proj}(\mathrm{v\_proj}(0))$,
i.e.\ the biases) belongs to the trainable conditioning, so training moves the no-prompt policy.
We pin it to its pretrained value (``frozen dummy'').
\end{itemize}

\subsection{The first full run and why pooled conditioning was not enough}\label{sec:sft-diag}
\paragraph{The first full run (pooled conditioning).}
We first trained the pooled recipe on all 30.1M sides with nine labels each: five LLM descriptions plus four templated
metadata labels (rating, time control, opening and trading level) at weight 0.05.
The text vector was Qwen3-Embedding-0.6B, last-token pooled and cut to its first 256 dimensions.
The model was trained Matryoshka-style \citep{kusupati2022matryoshka}, so any prefix of its 1,024-dimensional output is
itself a usable embedding; a linear layer maps the vector to the decoder width of 960.
We chose 256 dimensions because in early 10-minute runs the precomputed 256-dimensional vectors trained 51\% faster and
reached a lower cross-entropy than the full vector encoded online (2.9409 against 2.9507).
That comparison mixed two things, storage and width.
A clean rerun with both widths precomputed and stored alike (2M games each, same steps) found the full 1,024 dimensions
better: LLM-label cross-entropy 2.862 against 2.868 and a right-minus-wrong gap about 20\% larger ($-0.042$ against
$-0.036$), better in all five bands, at the same throughput.
Both were then superseded by per-token keys (below).
Top-player games were mixed in, rising from 1\% to 9.5\% of games over the run.
Training took one pass, 28,836 steps, on 8 GH200.
On the test set, cross-entropy is 2.880 with the right label, 2.929 with a wrong label and 2.905 with none.
The right-minus-wrong gap is $-0.049$, against $-0.303$ for the final model.
In play, concrete prompts change behavior (\cref{tab:sft-pooled-play}).
Requests for strong or weak play do not change strength: every training game came from a player rated $\ge2200$, so no
description was ever paired with weak play.
Player prompts barely work ($-0.002$ right vs.\ wrong player).
\end{verified}
\begin{table}[H]
\centering\small
\begin{tabular}{lll}
\toprule
prompt & behavior & no prompt \\
\midrule
\emph{``leave your king in the center and go for an open fight''} & castles in 45\% of games & 94\% \\
\pc & 39\% of castling queenside & 10\% \\
\pq & queens off by ply 40 in 40\% & 25\% \\
\pp & 0.289 pawn moves per move & 0.248 \\
strong-play prompts (8) / weak-play prompts (8) & 1765--1836 / 1778--1839 Elo & 1828 Elo \\
\bottomrule
\end{tabular}
\caption{The full pooled-conditioning run in play (200--300 games per prompt, 1,000 without a prompt).}
\label{tab:sft-pooled-play}
\end{table}
\begin{verified}{sft-5}
\paragraph{One pooled key cannot select information.}
With a single context token, the cross-attention softmax is trivially 1.
Each conditioned layer then just adds a text-dependent vector that is the same at every position.
To test whether this limited the first full run, we trained eight encoder variants on the same 500k games (\cref{tab:pooling}).
Passing \emph{all} token states as keys, with the same encoder, lowered cross-entropy by 0.023 and doubled the
right--wrong gap.
It also made fine details usable: the gap for an opening request (``play the French'') grew from $-0.006$ to $-0.070$.
The effect replicated on a second seed (seed noise is 0.001--0.002).
Among pooled encoders, every choice was within 0.003.
EmbeddingGemma-300m \citep{vera2025embeddinggemma} with per-token keys was best, and cheaper than Qwen with per-token
keys (\cref{fig:sft-enc}).
\end{verified}

\begin{table}[H]
\centering\small
\resizebox{\linewidth}{!}{\begin{tabular}{lrrrr}
\toprule
encoder and conditioning & LLM-label CE & mean gap & opening-template gap & relative time \\
\midrule
Qwen3-Emb-0.6B, pooled (early recipe) & 2.9137 & $-0.050$ & $-0.006$ & 1.00 \\
Qwen3-Emb-0.6B, pooled + instruction & 2.9109--2.9125 & $-0.052$ & $-0.006$ & 1.01 \\
EmbeddingGemma-300m, pooled & 2.9114 & $-0.054$ & $-0.007$ & 1.10 \\
Qwen3-Emb-4B, pooled + instruction & 2.9116 & $-0.055$ & $-0.003$ & 1.06 \\
Qwen3-Emb-0.6B, per-token & 2.8904 & $-0.097$ & $-0.070$ & 1.65 \\
Qwen3-Emb-0.6B, per-token, 3 conditioned layers & 2.8983 & $-0.086$ & $-0.055$ & 1.57 \\
\textbf{EmbeddingGemma-300m, per-token} & \textbf{2.8854} & $\mathbf{-0.100}$ & $\mathbf{-0.088}$ & 1.39 \\
\bottomrule
\end{tabular}}
\caption{Encoder comparison on identical data (500k games, 1,000 validation games). Gap = right minus wrong label CE.}
\label{tab:pooling}
\end{table}
\begin{verified}{sft-6}
\paragraph{Does the text vector keep a player's name?}
Before training, we tested whether a single text vector even keeps the details a user may write
(\cref{tab:probe}).
Prompts were rendered from our templates for 99 player names and their aliases, either alone (\emph{bare}) or appended to
a real medium-length description (\emph{mixed}, like ``aggressive, like Magnus''); the score is how often the nearest
neighbor rendered with a \emph{different} template names the same player (chance 1\%).
Alone, every embedder keeps the name.
Inside a sentence, pooled vectors lose it: Qwen3-Embedding-0.6B, last-token pooled as in our early recipe, finds the right
player in only 53\% of cases, and mean-pooled in 4.5\%.
EmbeddingGemma-300m keeps it best without an instruction (79\%); an instruction prefix naming the attributes recovers 99\%
for Qwen.
Per-token conditioning avoids the question altogether, since the decoder attends to every token, including the name.
Together with a larger share of player games, it raised the right--wrong player gap from $-0.002$ to $-0.103$
(\cref{sec:sft-final}).
\end{verified}
\begin{table}[H]
\centering\small
\begin{tabular}{lrr}
\toprule
encoder, pooling & bare & mixed \\
\midrule
Qwen3-Embedding-0.6B, last token (early recipe) & 98\% & 53\% \\
Qwen3-Embedding-0.6B, mean & 99\% & 4.5\% \\
Qwen3-Embedding-0.6B, last token, instruction & 100\% & 99\% \\
Qwen3-Embedding-8B, last token & 98\% & 57\% \\
EmbeddingGemma-300m, mean & 93\% & 79\% \\
Qwen3-0.6B language model, mean & 85\% & 6\% \\
\bottomrule
\end{tabular}
\caption{Player identification from one pooled text vector: share of prompts whose nearest neighbor, written with a
different template, names the same player (99 players, chance 1\%). No training involved.}
\label{tab:probe}
\end{table}

\section{Preference optimization}\label{sec:dpo}
\begin{verified}{dpo-1}
\subsection{Label-swap pairs}
Let $y$ be a real game, $c^+$ its own description and $c^-$ a description of a different game.
The \emph{chosen} sample is $(y,c^+)$ and the \emph{rejected} sample is $(y,c^-)$: the same moves, so only the text can
explain the preference.
No moves are sampled and no reward model is needed.
With the frozen SFT model as reference and $\Delta_\theta=\log\pi_\theta-\log\pi_{\text{ref}}$ summed over the described
side's moves, the loss is
\begin{align*}
\mathcal{L} &= -\log\sigma\!\big(\beta[\Delta_\theta(y|c^+)-\Delta_\theta(y|c^-)]\big)
+ \lambda_{\text{NLL}}\,\mathcal{L}_{\text{NLL}}(y|c^+)\\
&\quad+ \lambda_{\text{KL}}\sum_t \mathrm{KL}\big(\pi_{\text{ref}}(\cdot|y_{<t},c^+)\,\|\,\pi_\theta(\cdot|y_{<t},c^+)\big).
\end{align*}
An alternative, \emph{game pairs}, compares two different games under the same description; ``both'' uses both kinds.

\paragraph{Cost per game.}
Let $L_b=6$ be the number of decoder layers below the first conditioned layer and $L_t=6$ the number of conditioned
layers above it.
The bottom layers never see the text, so they run once per position and their output is shared by every description of
the game (\cref{sec:sft}); the top layers run once per position and description.
With $K$ descriptions per game, one training position therefore costs $L_b+K L_t$ layer applications:
\begin{itemize}\setlength\itemsep{0pt}
\item \textbf{SFT} conditions each game on all five bands, $K=5$: $6+5\cdot6=36$ layer applications per position,
forward and backward.
\item \textbf{DPO} draws one band uniformly at random per game and step and scores the game under the chosen description
(its own label in that band) and the rejected one (another game's label in the same band), $K=2$: $6+2\cdot6=18$ per
position, forward and backward, plus the same 18 for the frozen reference model, forward only.
\end{itemize}
Counting a backward pass as two forward passes, a position costs about $3\cdot36=108$ forward layer applications in SFT
and $3\cdot18+18=72$ in DPO, a predicted ratio of 1.5.
Measured on 8 GH200, SFT trained at about 2,400 games/s (300 per GPU) and the full-pass DPO at about 3,750 games/s
(470 per GPU), a ratio of 1.56.
Per description the order is reversed: SFT learns from five descriptions per game, about 12,000 per second, and DPO from
one, about 3,750 per second.

\subsection{Early rounds on the pooled model}
Six short rounds on the pooled-conditioning SFT model (\cref{sec:sft-diag}; 20 minutes per variant, 300 games per prompt) established the basic recipe:
\begin{itemize}\setlength\itemsep{1pt}
\item \textbf{DPO changes play.} Under \pc, queenside castling rose from 39\% to 72--87\%; queen trades under \pq rose
from 40\% to 48--57\%, and pawn moves increased.
No-prompt strength stayed within noise.
\item \textbf{Label pairs need an anchor.} With a weak NLL term, label-only pairs found a shortcut: make the
conditioning destructive, so that the game becomes unlikely under any wrong label.
Right-label cross-entropy rose by 0.34 and Elo fell by 50--145.
\item \textbf{The no-text path must be pinned.} Even ``conditioning-only'' DPO, with the decoder frozen, moved the
no-prompt policy through the cross-attention biases (\cref{sec:sft-diag}).
Pinning the whole empty-context term leaves the no-prompt model exactly unchanged.
\item \textbf{An NLL weight of 1.0} recovered 40--85 Elo under risky prompts at little cost in adherence.
\item \textbf{Queen trades need decoder updates.} Conditioning-only DPO moved castling and pawn play but not queen trades.
\end{itemize}

\subsection{Search on the final SFT model}
We ran about 70 short DPO runs (10--20 minutes on one GPU) in an \texttt{autoresearch}-style loop
\citep{karpathy2026autoresearch}.
A queue file held one configuration per line, workers claimed lines, and every run was logged with all its metrics.
Early runs were screened with 100 games per prompt; from round 6 on, with 300 games without a prompt, 300 under \pa
and 150 under each other style prompt.
The 20 most promising were re-evaluated at 1,000 games per prompt.

\paragraph{Variants.}
\begin{itemize}\setlength\itemsep{0pt}
\item pair type (label, game, both, single moves) and which bands form pairs;
\item $\beta$, NLL and KL weights; wins-only data; hard negatives (the most similar label in the batch);
\item a frozen chess body (conditioning-only) and different conditioning learning rates;
\item \emph{style-move DPO}: the preference uses only the top 10--50\% of moves where the reference model's
log-likelihood depends most on the label, and the KL or NLL anchor covers the rest;
\item IPO \citep{azar2024ipo}; weight averaging \citep{wortsman2022soups}; test-time guidance
$\log p=\log p_{\text{ref}}+w(\log p_\theta-\log p_{\text{ref}})$ \citep{ho2022cfg}.
\end{itemize}

\paragraph{What we saw at 1,000 games.}
\begin{itemize}\setlength\itemsep{1pt}
\item Every DPO variant moved all three style metrics well beyond SFT: queens 51\% $\to$ 62--71\%, pawns .292 $\to$
.309--.320, queenside 97\% $\to$ 98--100\%.
\item Every variant also enlarged the attack gap: SFT loses 41 Elo under \pa, and DPO models lose 91--185.
Guidance on SFT alone showed the same coupling: at $w{=}1.5$ it matched DPO's style with a gap of $-209$, and at
$w{=}3$ pawns rose to .365 at a gap of $-417$.
\item Style-move DPO gave the same style with about 50 Elo less loss under the attack prompt.
\item $\beta=1$, IPO, weaker KL, a higher body learning rate, removing the anchor and hard negatives all traded strength
for style without improving the trade-off.
\item Screening at 100 games misled: one run screened at a $-66$ gap and measured $-165$ at 1,000 games.
\end{itemize}

\paragraph{Evaluation and selection.}
No single number captures a style model, so each run is judged on a set of high-level play metrics: queen trades
under \pq, pawn moves per move under the pawn-storm prompt, queenside castling under \pc, Elo without a prompt, and the
attack gap (Elo under \pa minus Elo without a prompt).
The attack gap is not a metric we can rely on: it only signals that play under the risky prompt diverges from
no-prompt play, which we assume reflects real prompt following.
On that assumption we count a more negative gap as better, as one signal among five.
To rank runs on one scale, each metric is standardized against the short DPO runs ($z$ score) and the five are summed,
both unweighted and weighted (weights 1, 1, 0.2 for the saturated queenside metric, 1 and 1, adjusted for their
correlations; \cref{sec:z}).
\Cref{fig:dpo-search} shows the unweighted score of every screening run in launch order.
\Cref{fig:dpo-tradeoff} shows the trade-off behind the score: every DPO run follows the queen-trade prompt far more
often than SFT, at a larger attack gap.
\end{verified}
\begin{figure}[H]
\centering
\includegraphics[width=\linewidth]{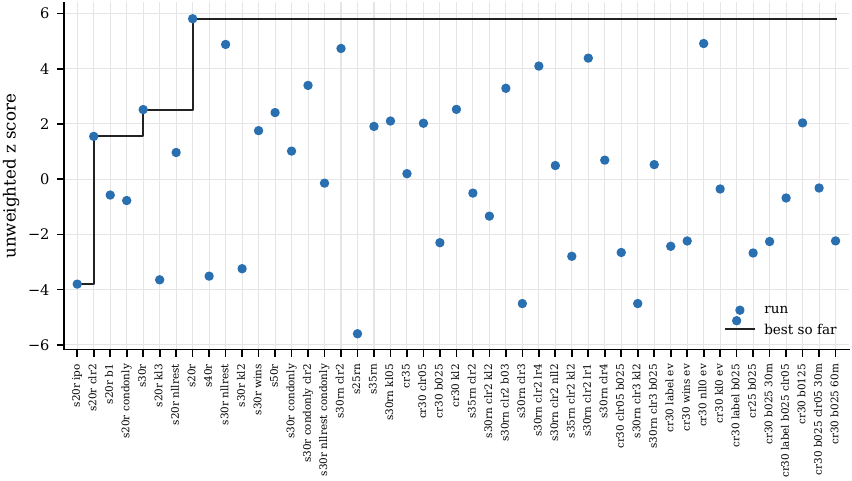}
\caption{DPO search: unweighted $z$ score of each run at screening, runs on the horizontal axis in launch order, with
the best so far; early rounds used 100 games per prompt, later ones 300 for no prompt and the attack prompt and 150
for the others. The $z$ scores are relative to the DPO runs themselves; any DPO run scores far above SFT (about $-20$,
\cref{fig:dpo-zsel}), so the score ranks DPO recipes, not DPO against SFT.}
\label{fig:dpo-search}
\end{figure}

\begin{figure}[H]
\centering
\begin{minipage}[c]{0.6\linewidth}
\includegraphics[width=\linewidth]{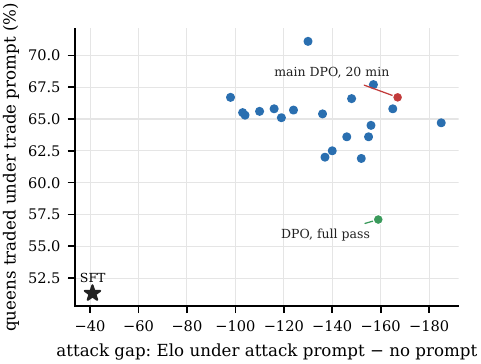}
\end{minipage}\hfill
\begin{minipage}[c]{0.36\linewidth}
\caption{Style adherence against the attack gap for SFT and the 20 DPO runs re-evaluated at 1,000 games per
prompt (blue). Red: the main 20-minute run; green: the full pass.}
\label{fig:dpo-tradeoff}
\end{minipage}
\end{figure}

\begin{figure}[H]
\centering
\includegraphics[width=\linewidth]{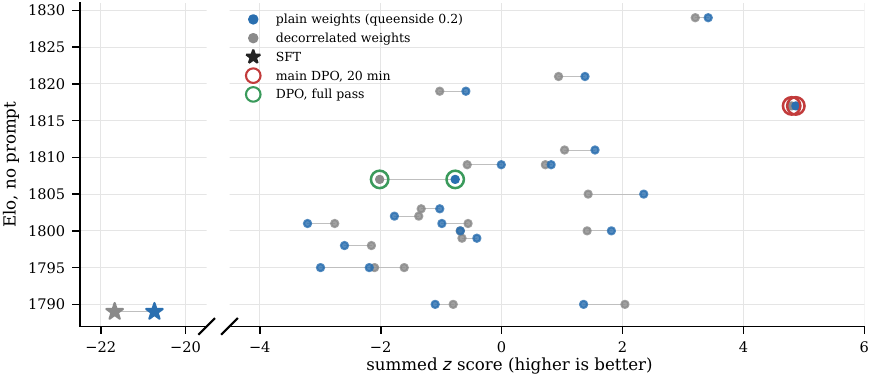}
\caption{Selection of the main model: summed $z$ score against no-prompt Elo for the 20 short DPO runs, the full pass and SFT
(star), all at 1,000 games per prompt, each scored twice: blue with weight 0.2 on the saturated
queenside metric and 1 elsewhere, gray with the decorrelated weights of \cref{sec:z}. Red circles: the selected 20-minute
run; green circles: the full pass.}
\label{fig:dpo-zsel}
\end{figure}
\begin{figure}[H]
\centering
\includegraphics[width=\linewidth]{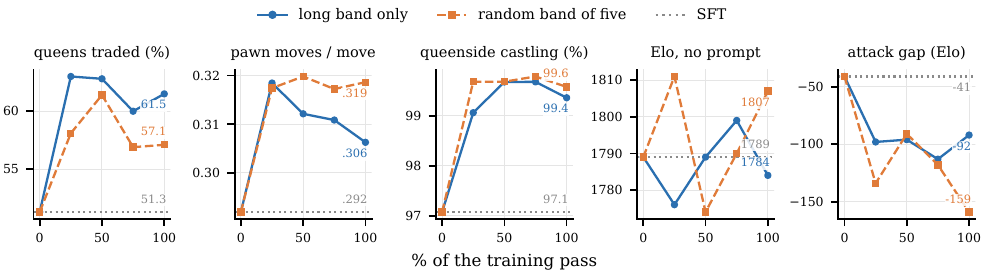}
\caption{Full-pass DPO with pairs from the long band only (solid) vs.\ a random band of five (dashed), one panel per
metric; 1,000 games per point; the dotted line is SFT.}
\label{fig:curve}
\end{figure}

\begin{verified}{dpo-2}
\paragraph{The main model.}
The top run used label pairs, $\beta=0.5$, NLL 1.0, KL 1.0 on all moves, a pinned no-text term, body learning rate
$2{\times}10^{-7}$ and conditioning learning rate $10^{-4}$: 20 minutes on one GPU, 4,366 steps of 128 games, about
0.56M games or 2\% of the corpus.
Because it was the best of about 70 noisy runs, its lead could be a lucky initialization or an artifact of tuning to the
evaluation metrics (\cref{sec:noise}).
We therefore re-evaluated it on a second, independent set of 1,000 games per prompt.
On 1,000 games per prompt it had the highest score under both weightings, $z=+4.9$ and $+4.8$, against $-0.8$ and
$-2.0$ for the full pass and $-20.7$ and $-21.7$ for SFT (\cref{fig:dpo-zsel,tab:bs}).
Pooled with the second evaluation, 2,000 games per prompt, its score falls to $+2.6$ and $+2.8$ (full pass $-0.4$ and
$-1.4$, SFT $-20.3$ and $-21.0$), part of the first lead being luck, but it keeps its adherence and its no-prompt
strength (\cref{tab:headline}), so we release this 20-minute run as the main DPO model (\texttt{model.pt}).
We also trained the same recipe for one full pass over the corpus (below) and release that model, with its
intermediate checkpoints, for others to test.

\subsection{Scaling up, pair construction and batch size}
\paragraph{Full pass.}
We ran the selected recipe for one full pass over 27.08M games: 8 GPUs $\times$ 256 games, 13,428 steps, 2.0 hours.
It did \emph{not} improve on the 20-minute run (\cref{fig:curve}).
Adherence saturates within the first quarter, while validation pair accuracy keeps rising (0.89 $\to$ 0.93) and
wrong-label cross-entropy keeps growing (3.60 $\to$ 4.08).
Longer DPO sharpens label discrimination, not play.
\end{verified}
\begin{verified}{dpo-3}
\paragraph{Pair construction.}
In 15-minute runs (\cref{tab:pairs}), using one random band, the long band or all five bands made no reliable
difference, and all five bands was 4$\times$ slower.
Preferences per move instead of per game gave a gentler regime: an attack gap of only $-55$ to $-74$, close to SFT, but
little extra style.
A full long-band-only pass is shown in \cref{fig:curve}; two identical long-band runs ended with
gaps of $-143$ and $-92$.
We count the full passes as a win for the random band: at the end of the pass it keeps more pawn play (.317 against
.308) and more no-prompt strength (1812 against 1798 Elo, 2,000 games), and its summed $z$ is $-0.4$ against $-5.6$;
the long band keeps slightly more queen trades (60.9\% against 58.7\%).
\end{verified}

\begin{verified}{dpo-4}
\paragraph{Batch size.}
Why did the full pass not beat the 20-minute run?
A controlled test isolates batch size: same recipe, 20 minutes each (\cref{tab:bs}).
Queen trades fall as the batch grows, while 4 GPUs $\times$ 32 games matches 1 GPU $\times$ 128 games, which rules out
the multi-GPU code path.
We first suspected the step size: AdamW \citep{loshchilov2019adamw} moves the weights by roughly one learning rate per
step regardless of batch size, so at a fixed learning rate a larger batch moves the weights less per game seen
\citep{goyal2017largebatch}.
Run D tested this with the learning rate scaled by 2.8 and did not recover the style, so this explanation is not
supported.
Batch size and the number of games seen are confounded here (C and D saw four times more games), and the differences
are only 1--2 standard errors, so the cause remains open.
\end{verified}

\clearpage
\begin{table}[H]
\centering\footnotesize\setlength\tabcolsep{4pt}
\begin{tabular}{@{}llrrrrr@{}}
\toprule
bands & granularity & queens (\%) & pawns & queenside (\%) & Elo, none & attack gap \\
\midrule
SFT & --- & 51.3 & .292 & 97.1 & 1789 & $-41$ \\
\midrule
one random & game & 65.5 & .317 & 99.1 & 1788 & $-157$ \\
long only & game & \textbf{67.4} & \textbf{.321} & 99.1 & \textbf{1811} & $-150$ \\
all five & game & 65.2 & .315 & 99.2 & 1793 & $-160$ \\
all five, label + game pairs & game & 65.3 & .315 & \textbf{99.4} & 1806 & $-168$ \\
one random & move & 56.9 & .298 & 99.1 & 1792 & $-74$ \\
long only & move & 58.7 & .297 & 98.9 & 1768 & $-59$ \\
all five & move & 58.0 & .292 & 98.0 & 1787 & $-55$ \\
\bottomrule
\end{tabular}
\caption{Pair construction, 15-minute runs, 1,000 games per prompt.}
\label{tab:pairs}
\end{table}

\begin{table}[H]
\centering\small
\resizebox{\linewidth}{!}{\begin{tabular}{llrrrrrrrr}
\toprule
run & GPUs $\times$ games & games seen & queens (\%) & pawns & queenside (\%) & Elo, none & attack gap & $z$ & $z$, decorr. \\
\midrule
main & 1 $\times$ 128 & 0.56M & \textbf{66.7} & .318 & 99.7 & \textbf{1817} & $-167$ & $\mathbf{+4.9}$ & $\mathbf{+4.8}$ \\
C & 4 $\times$ 256 & 2.53M & 61.6 & .318 & 99.4 & 1809 & $-157$ & $+1.1$ & $+0.4$ \\
A, new seed & 1 $\times$ 128 & 0.59M & 64.8 & .315 & 98.1 & 1810 & $-139$ & $+0.3$ & $-0.1$ \\
D, learning rate $\times$2.8 & 4 $\times$ 256 & 2.50M & 61.0 & .317 & \textbf{99.8} & 1796 & $-152$ & $-0.6$ & $-1.1$ \\
B & 4 $\times$ 32 & 0.59M & 63.1 & .316 & 99.5 & 1795 & $-140$ & $-0.7$ & $-0.8$ \\
full pass & 8 $\times$ 256 & 27.08M & 57.1 & \textbf{.319} & 99.6 & 1807 & $-159$ & $-0.8$ & $-2.0$ \\
\midrule
SFT & --- & --- & 51.3 & .292 & 97.1 & 1789 & $-41$ & $-20.7$ & $-21.7$ \\
\bottomrule
\end{tabular}}
\caption{Batch-size test: identical recipe and 20-minute budget except the full pass; 1,000 games per prompt for every
row (the first evaluation of the main run and of the full pass), sorted by $z$. $z$: summed $z$ score against the 20 short DPO runs, weight 0.2 on queenside and 1 elsewhere; decorr.: the
decorrelated weights of \cref{sec:z}.}
\label{tab:bs}
\end{table}

\section{Final results}\label{sec:results}
\begin{verified}{results-1}
\Cref{tab:headline} compares SFT with the released DPO models at 2,000 games per prompt (two independent 1,000-game
evaluations pooled).
Validation likelihoods are in \cref{tab:valce}.
\end{verified}
\begin{table}[H]
\centering\small
\begin{tabular}{lrrrrrr}
\toprule
 & \multicolumn{3}{c}{style adherence} & \multicolumn{2}{c}{strength} & \\
\cmidrule(lr){2-4}\cmidrule(lr){5-6}
model & queens (\%) & pawns & queenside (\%) & Elo, none & attack gap & $z$ \\
\midrule
SFT (start) & 53.0 & .290 & 97.3 & 1792 & $-40$ & $-$21.0 \\
\midrule
DPO, 20-min run (main) & \textbf{66.5} & \textbf{.319} & \textbf{99.7} & 1797 & $-150$ & \textbf{+2.8} \\
DPO, full pass & 58.7 & .317 & 99.6 & 1812 & $-155$ & $-$1.4 \\
DPO, long band only, 25\% & 64.2 & .318 & 99.1 & 1790 & $-107$ & $-$1.1 \\
DPO, long band only, 100\% & 60.9 & .308 & 99.6 & 1798 & $-113$ & $-$5.6 \\
\bottomrule
\end{tabular}

\caption{Main results, 2,000 games per prompt. Bold marks the best value per column.
95\% precision is about $\pm$1--2 points for shares and $\pm$21 Elo.}
\label{tab:headline}
\end{table}

\begin{table}[H]
\centering\small
\begin{minipage}[c]{0.65\linewidth}
\centering
\begin{tabular}{lrrrr}
\toprule
model & right & wrong & none & pair accuracy \\
\midrule
SFT & \textbf{2.806} & 3.094 & \textbf{2.928} & 0.803 \\
DPO, 20-min run (main) & 2.846 & 3.600 & 2.932 & 0.893 \\
DPO, full pass & 2.851 & \textbf{4.077} & 2.933 & \textbf{0.925} \\
\bottomrule
\end{tabular}
\end{minipage}\hfill
\begin{minipage}[c]{0.32\linewidth}
\caption{Validation cross-entropy (nats per move, described side) with the right, a wrong or no description.
Pair accuracy is the share of games whose moves are more likely under their own description than under another's.}
\label{tab:valce}
\end{minipage}
\end{table}

\begin{figure}[H]
\centering
\includegraphics[width=\linewidth]{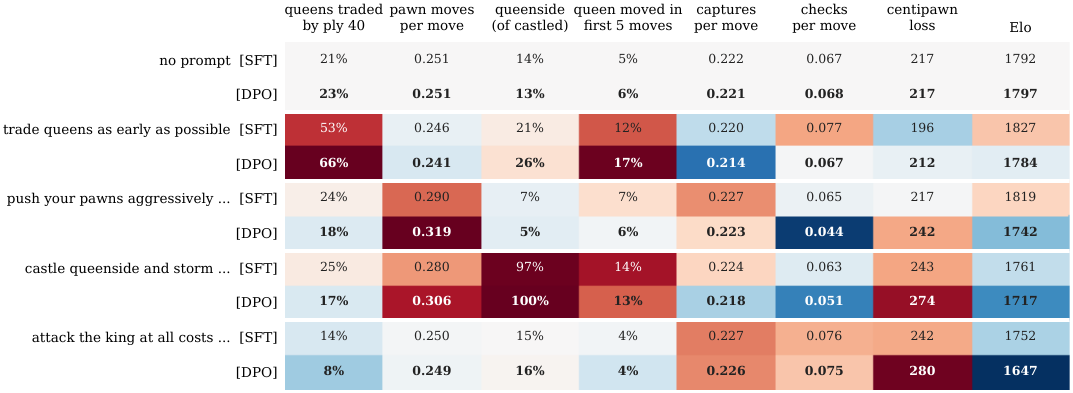}
\caption{Behavior of the model's own side under the four main prompts, SFT against the main DPO model (bold), 2,000
games per cell. Cell text is the value; color is the change from the same model with no prompt, in units of the
column's spread (red: higher, blue: lower). Under each prompt, DPO moves the targeted behavior further than SFT: more
early queen trades, more pawn moves, more queenside castling, and a much larger loss of strength under the attack prompt.}
\label{fig:style-heatmap}
\end{figure}

\begin{table}[H]
\centering\footnotesize\setlength\tabcolsep{4pt}
\resizebox{\linewidth}{!}{\begin{tabular}{@{}llccc@{}}
\toprule
prompt $\rightarrow$ metric & model & no prompt & right prompt & other prompts \\
\midrule
\pq $\rightarrow$ queens traded by ply 40 & SFT & $0.21\pm0.41$ & $0.53\pm0.50$ & $0.21\pm0.41$ \\
 & DPO & $0.23\pm0.42$ & $\mathbf{0.66\pm0.47}$ & $0.14\pm0.35$ \\
\pp $\rightarrow$ pawn moves per own move & SFT & $0.251\pm0.072$ & $0.290\pm0.076$ & $0.259\pm0.075$ \\
 & DPO & $0.251\pm0.074$ & $\mathbf{0.319\pm0.074}$ & $0.265\pm0.077$ \\
\pc $\rightarrow$ queenside castling, all games & SFT & $0.13\pm0.34$ & $0.89\pm0.31$ & $0.13\pm0.33$ \\
 & DPO & $0.13\pm0.33$ & $\mathbf{0.94\pm0.25}$ & $0.13\pm0.34$ \\
\pa $\rightarrow$ checks per own move & SFT & $0.067\pm0.071$ & $0.076\pm0.073$ & $0.068\pm0.073$ \\
 & DPO & $0.068\pm0.074$ & $0.075\pm0.071$ & $0.054\pm0.062$ \\
\bottomrule
\end{tabular}}
\caption{Targeted behavior under the matching prompt, under no prompt and under the other three main prompts: mean
$\pm$ standard deviation over games, all against Stockfish at UCI\_Elo 2000. $n=2{,}000$ games for no prompt and the right
prompt, $6{,}000$ for the other prompts. \Cref{fig:style-bars} repeats this in self-play with random descriptions from
the dataset as the baseline.}
\label{tab:rightwrong}
\end{table}

\begin{figure}[H]
\centering
\includegraphics[width=\linewidth]{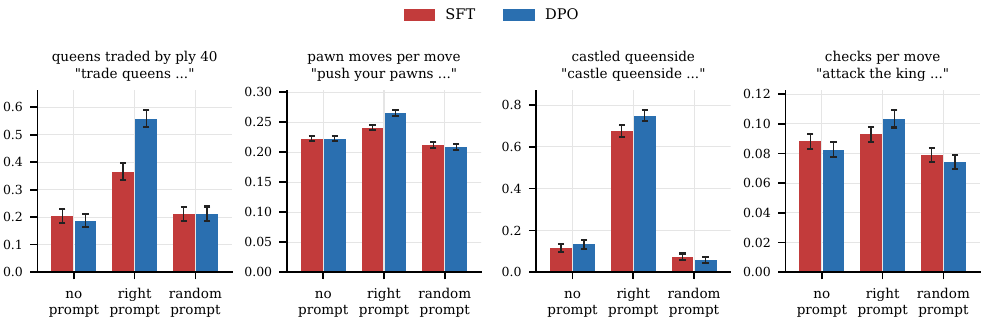}
\caption{Self-play style evaluation (\cref{sec:eval}): the targeted behavior of the prompted side for SFT and DPO with
no prompt, the right prompt and a random test-split description (a different one per game), 1,000 games per bar
against \textsc{chsm8-player-base}; bars: 95\% intervals over games.}
\label{fig:style-bars}
\end{figure}
\begin{figure}[H]
\centering
\includegraphics[width=\linewidth]{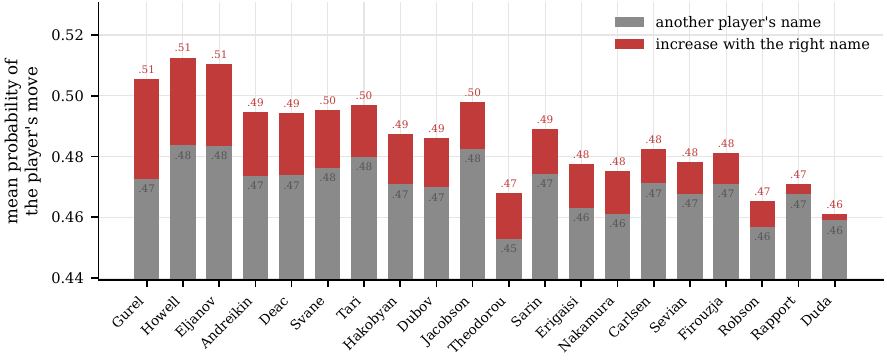}
\caption{Player prompts in \dpo, paired on about 1,000 held-out games per player: mean probability, over legal moves,
of the player's actual move under another top player's name (gray) and its increase with the right name (red). The
vertical axis starts at 0.44.}
\label{fig:player-prompts}
\end{figure}

\begin{verified}{results-2}
\paragraph{Player prompts in the final model.}
To test whether \emph{``play like \{player\}''} makes \dpo play more like that player, we took the 20 top players with
the most held-out games (about 1,000 each, 20,508 games and 925k of the players' own moves, none seen in training) and
scored every game twice: once with the player's name and once with the same wording and another top player's name.
Pairing on the same game removes the game-to-game variation.
With the right name, the mean probability of the player's actual move rises from 47.1\% to 48.7\%, top-1 move match
from 53.7\% to 55.5\% and top-3 match from 77.4\% to 78.8\%, and it rises for all 20 players
(\cref{fig:player-prompts}).
\end{verified}
\begin{verified}{results-3}
\begin{itemize}\setlength\itemsep{1pt}
\item \textbf{DPO clearly improves prompt following} (\cref{tab:headline,fig:style-heatmap}).
Every DPO model improves every adherence metric over SFT.
In $z$ units SFT sits at $-21$, while all DPO settings fall within roughly $0\pm2$.
DPO mostly pushes wrong descriptions down (3.09 $\to$ 3.60--4.08 nats) and barely moves right ones.
\item \textbf{SFT already follows prompts; DPO sharpens it.}
SFT alone moves every targeted behavior well away from its no-prompt play, for example queenside castling from 13\% to
89\% of all games under \pc (\cref{tab:rightwrong}).
DPO adds to an already tuned model: 89\% $\to$ 94\% there, and larger gains where SFT was weaker (queens traded by ply 40
under \pq, 53\% $\to$ 66\%).
Under a prompt that asks for something else, both models play close to their no-prompt baseline, and DPO moves
slightly away from the unrequested behavior (queens traded 23\% $\to$ 14\%).
In self-play against random descriptions from the dataset the picture is the same (\cref{fig:style-bars}): random
descriptions leave each behavior at or below its no-prompt level, while the right prompt moves it, and DPO moves it
further, including checks under \pa, which barely moved against Stockfish.
\item \textbf{Strength without a prompt is unchanged}: 1790--1812 Elo against 1792 for SFT.
\item \textbf{Faithful prompt following costs strength under risky prompts.}
Under \pa, DPO models lose about 150 Elo relative to no prompt, against 40 for SFT.
Sacrificial training games were not lost more often (53.8\% wins): speculative sacrifices against an accurate engine
without search simply lose.
\end{itemize}

\subsection{Against strong public models}\label{sec:versus1k}
We extended the head-to-head comparison of \cref{sec:pt-results} to 1,000 games per match and to recent strong models
(\cref{tab:versus1k}): \pt, \textsc{chsm8-player-base} and the Carlsen fine-tune each played every opponent, both
sides choosing their most likely legal move from book openings played with both colors.
To place every opponent on our scale, each also played 1,000 games against Stockfish at \texttt{UCI\_Elo}~2000 under
our Elo protocol (\cref{sec:eval}).
ChessLFM is played twice: with its one-pass policy, and with its own depth-3 search, which we list separately.
Human-move models are given a rating of 2800 for both players where they take one (Otter: its top bucket, 2000+).
We chose the Carlsen fine-tune simply because we wanted one player model in the comparison and its checkpoint was among
our best; we did not expect it to beat the strong recent models.
\begin{itemize}\setlength\itemsep{1pt}
\item \textbf{Older public models}: all three \chsm models win every match, by 65--94\%; \textsc{chsm8-player-base},
trained on stronger games, wins each one by more than \pt.
\item \textbf{Strong recent models without search}: \chsm is about even with Otter (49--61\%) and loses clearly to
Allie 2.0, Maia-3 and DeepMind's searchless transformers (6--21\%), which rate 2230--2580 against Stockfish where our
three models rate 1837--1903.
These models are trained on far more data or on engine evaluations, and none of them takes a text prompt.
C1-4B, a language model that reasons about the position in text, is weak as a player: it was trained to solve chess
problems, not to play whole games, and over 1,000 games against Stockfish it scored 12 draws and no wins (about 1110 Elo);
\pt scores 98.6\% against it.
QUEEN (PAWN-8), also a reasoning language model, is much stronger (2354 Elo over 200 games) and wins 81--86\% of 200
games against our models.
In both settings 6--11\% of its answers were missing or illegal and were re-decoded restricted to legal moves.
\item \textbf{With search}: ChessLFM \citep{labonne2026chesslfm} gains about 310 Elo from its own search (1687
$\rightarrow$ 2001) at 30 times the time per move (24 $\rightarrow$ 759\,ms). Our models score 91--93\% against it without
search and 48--60\% with it: \pt is about even, and the two games-trained models stay ahead by about 50--70 Elo with
one forward pass per move. Matilda, which re-ranks Maia-3's candidates with Stockfish, rates above 3000 and wins
almost every game.
\end{itemize}
\end{verified}
\begin{table}[H]
\centering\footnotesize\setlength\tabcolsep{3pt}\renewcommand\arraystretch{0.92}
\begin{tabular}{@{}lrrrrrr@{}}
\toprule
opponent & params & Elo (95\% CI) & ms/move & \pt & \textsc{player-base} & \textsc{player-carlsen} \\
\midrule
\multicolumn{7}{@{}l}{\emph{public models trained by supervised prediction}} \\
\href{https://huggingface.co/shermansiu/maia2-blitz}{Maia-2 blitz} & 23M & 1657 (1626--1684) & 3 & 65.3\% & \textbf{71.8\%} & 71.7\% \\
\href{https://huggingface.co/mlabonne/LFM2.5-230M-Chess}{ChessLFM} & 231M & 1687 (1659--1713) & 24 & 91.5\% & \textbf{93.3\%} & 92.8\% \\
\href{https://huggingface.co/Waterhorse/chessgpt-base-v1}{ChessGPT-base} & 2.8B & 1731 (1708--1752) & 67 & 88.0\% & 93.7\% & \textbf{94.3\%} \\
\href{https://huggingface.co/datasets/ezipe/lichess-models/blob/main/50M-High-Elo-1500-No-Elo-Conditioning/2024-04-30---03-07-10_50M-High-Elo-1500-No-Elo-Conditioning/ckpt_1030000.pt}{ChessFormer (Transcendence)}\footnotemark & 51M & 1733 (1710--1754) & 39 & 86.5\% & 90.0\% & \textbf{93.9\%} \\
\href{https://huggingface.co/austindavis/chess-gpt2-uci-12x12x768}{chess-gpt2-uci} & 86M & 1784 (1763--1803) & 24 & 82.2\% & 88.8\% & \textbf{91.0\%} \\
\href{https://huggingface.co/adamkarvonen/chess_llms}{Chess-GPT (Karvonen)} & 51M & 1873 (1853--1892) & 40 & 75.1\% & \textbf{83.3\%} & 81.1\% \\
\midrule
\multicolumn{7}{@{}l}{\emph{strong recent models, no search}} \\
\href{https://huggingface.co/UofTCSSLab/C1-4B}{C1 (reasons in text)} & 4.0B & 1112 (968--1190) & 570 & 98.6\% & 99.2\% & \textbf{99.4\%} \\
\href{https://huggingface.co/peargentlabs/otter-chess}{Otter} & 15M & 2021 (2002--2041) & 7 & 49.1\% & 56.1\% & \textbf{61.1\%} \\
\href{https://huggingface.co/yimingzhang/allie-2.0}{Allie 2.0} & 5.6B & 2232 (2210--2256) & 31 & 13.5\% & \textbf{21.2\%} & 19.8\% \\
\href{https://huggingface.co/UofTCSSLab/Maia3-79M}{Maia-3} & 79M & 2271 (2247--2296) & 13 & 14.0\% & 15.7\% & \textbf{21.3\%} \\
\href{https://huggingface.co/princeton-nlp/queen_pawn-8}{QUEEN (PAWN-8), 200 games} & 3.3B & 2354 (2304--2420) & 35{,}227 & 14.2\% & 17.2\% & \textbf{19.0\%} \\
\href{https://github.com/google-deepmind/searchless_chess}{Searchless} & 9M & 2403 (2377--2433) & 8 & 14.9\% & 13.0\% & \textbf{16.7\%} \\
\href{https://github.com/google-deepmind/searchless_chess}{Searchless} & 270M & 2505 (2473--2542) & 23 & 8.0\% & 7.1\% & \textbf{12.2\%} \\
\href{https://github.com/google-deepmind/searchless_chess}{Searchless} & 136M & 2584 (2546--2632) & 14 & 10.2\% & \textbf{10.3\%} & 6.3\% \\
\midrule
\multicolumn{7}{@{}l}{\emph{with search}} \\
\href{https://huggingface.co/mlabonne/LFM2.5-230M-Chess}{ChessLFM with search} & 231M & 2001 (1983--2020) & 759 & 47.7\% & 57.5\% & \textbf{59.7\%} \\
\href{https://github.com/GarryChess/matilda-uci}{Matilda (Stockfish depth 12)} & 79M + SF & 3009 (2906--3287) & 495 & 1.9\% & \textbf{4.1\%} & 1.6\% \\
\bottomrule
\end{tabular}

\caption{1,000-game matches: score of each \chsm model against each public model (wins plus half the draws; bold:
best of ours) and its Elo against Stockfish at \texttt{UCI\_Elo}~2000 (ours: 1837, 1890, 1903). Names link to the exact
weights we ran; params of those weights (ours: 210.5M, 166.2M); ms/move at batch size 1 on one GH200 (ours: 13--14).}
\label{tab:versus1k}
\end{table}
\footnotetext{\raggedright A dataset repository holding the checkpoints; we used
\path{50M-High-Elo-1500-No-Elo-Conditioning/2024-04-30---03-07-10_50M-High-Elo-1500-No-Elo-Conditioning/ckpt_1030000.pt}.}

\noindent Against Stockfish alone, a proxy for Elo, our models rate below ChessLFM with search (2001) and Otter (2021),
yet both lose to \textsc{player-base} (1890) and \textsc{player-carlsen} (1903).
A Bradley--Terry fit \citep{bradley1952rank} to all 64,746 games uses every head-to-head result instead of one engine
and ranks our models above the opponents they beat (\cref{app:bt}).
Lc0 and Maia-1 could not be run: the Lc0 engine they share aborted repeatedly on our nodes.
Otter can be tuned in strength through both players' rating, the time control and the clock; we used its strongest
rating bucket for both sides and its defaults for the rest (600+0, half the clock left).

\begin{verified}{results-4}
\subsection{How noisy is this?}\label{sec:noise}
At 1,000 games per prompt, the measurement noise is about as large as the spread between DPO recipes:
\end{verified}
\begin{center}\small
\begin{tabular}{lrrrrr}
\toprule
 & queens (pts) & pawns & queenside (pts) & Elo, none & attack gap \\
\midrule
spread between 20 recipes (std.) & 2.1 & .0029 & 0.38 & 10 & 24 \\
noise of one 1,000-game eval & 1.5 & .0016 & 0.33 & $\approx$15 & $\approx$21 \\
\bottomrule
\end{tabular}
\end{center}\begin{verified}{results-5}
\noindent Only queens and pawns separate recipes.
The same main checkpoint scored 1817 Elo and $z=+4.8$ on one 1,000-game batch, and 1776 Elo and $z=+0.8$ on another.
Because the main model was the best of about 70 noisy runs, part of its lead is likely luck (the winner's curse
\citep{thaler1988winners}), or the result of tuning to our five evaluation prompts rather than a generally better model.
A retrain with a new seed reached 64.8\% queens rather than 66.5\%.

\subsection{Selection score}\label{sec:z}
For each metric $m$ we compute $z_m=(x_m-\mu_m)/s_m$ against the 20 short DPO runs of the 1,000-game re-evaluation,
oriented so that higher is better.
The plain score is $\sum_m w_m z_m$ with $w=(1,1,0.2,1,1)$.
The attack gap and no-prompt Elo correlate ($+0.60$), so no-prompt Elo would be counted twice.
The reported score therefore shrinks the correlation matrix $R$ 50\% toward the identity and uses weights $R^{-1}w$
rescaled to the same total: $(w_{\text{queens}}, w_{\text{pawns}}, w_{\text{queenside}}, w_{\text{Elo}},
w_{\text{gap}}) = (1.28, 1.02, 0.33, 0.99, 0.57)$.
A product of percentiles was rejected: it assumes independent columns and is dominated by the noisiest one.
\end{verified}

\section{Lessons and limitations}\label{sec:lessons}
\paragraph{What mattered most.}
\begin{itemize}\setlength\itemsep{1pt}
\item \textbf{Grounded labels.} A feature card measured from the board made LLM descriptions true and checkable.
A teacher that writes vivid but generic text (Apertus-v1.5-70B) carried 4--5$\times$ less signal, however much of it we had.
\item \textbf{Per-token conditioning.} The single largest SFT improvement came from passing all token states to
cross-attention instead of a pooled vector, not from a bigger encoder.
\item \textbf{Data needs fall sharply after pretraining.} Pretraining needed about a billion games.
SFT needed far less, tens of millions of described sides, and already followed style prompts well halfway through.
DPO needed very little, about half a million games, and on top of a good SFT model it is a modest increment.
\item \textbf{Simple preferences.} Label-swap DPO on human games gave large adherence gains without sampling, reward
models.
\end{itemize}

\paragraph{What we would do differently.}
\begin{itemize}\setlength\itemsep{1pt}
\item \textbf{Evaluate with more power.} 100--200 games per prompt are not representative (about $\pm50$ Elo): one run
screened at a $-66$ attack gap with 100 games and measured $-165$ at 1,000.
1,000 games per prompt is enough to rank recipes, 2,000 makes a final choice safe, and beyond that the returns
diminish.
\item \textbf{Tune at the target scale.} Learning rates and DPO settings chosen in short runs did not transfer to long
runs, both in SFT (short-horizon bias) and in DPO (the full pass did not beat a 20-minute run).
\item \textbf{Curricula.} Our pretraining order was a deliberate middle ground between a uniform shuffle, for general
play, and a bias toward modern games, which we wanted the model to imitate.
Whether that bias helped is not known; the resulting strength is good, but a curriculum chosen from tested evidence
rather than intuition would be the sounder choice.
\end{itemize}

\paragraph{Engineering.}
Several problems cost more time than any modeling decision.
\begin{itemize}\setlength\itemsep{1pt}
\item \textbf{Input pipeline.} Per-step bin-packing in Python left the GPUs idle, at 0\% utilization for minutes at a
time. Packing offline gave 3.5--4$\times$, and a background prefetch thread gave 9--12$\times$ on the full 1.4\,TB corpus.
\item \textbf{Storage.} Random reads from the shared Lustre filesystem were 11$\times$ slower than sequential reads, so
training data belongs on node-local disk or in RAM; in SFT, random per-game reads cost 43\,ms per game until the data
was held in RAM.
\item \textbf{Tokenization speed.} Converting games to move tokens with python-chess was slow; a Rust library such
as \texttt{shakmaty} would make it many times faster.
\item \textbf{The evaluation engine.} Stockfish at 0.05\,s per move took about 4 times longer per move than our model
even at batch size 1 (13\,ms), and far longer once our moves are batched across games, so evaluation was bound by the
CPU engine; a simpler, faster reference opponent would have made evaluation much cheaper.
\item \textbf{Compiled checkpoints.} A checkpoint saved from a \texttt{torch.compile}d model carried prefixed keys and
could not be loaded; models are now saved with clean keys.
\end{itemize}

\paragraph{Limitations.}
\begin{itemize}\setlength\itemsep{1pt}
\item \textbf{The alignment tax}: SFT costs about 45 Elo without a prompt (1792 vs.\ 1837), because the decoder is
trained to fit the description as well as the move.
\item \textbf{No weak-play conditioning}: every described player is rated $\ge2200$, so prompts asking for weak or strong
play have no effect on strength.
\item \textbf{Players}: player prompts work for 94 of 99 players, but the effect is modest (about one point of move
match), because top players differ little in their moves.
\item \textbf{Proxy metrics}: adherence is measured on a handful of prompts with hand-written metrics, not a validated
judgment of style.
\item \textbf{No search}: a little search \citep{zhang2025allie} could reduce the
strength cost of risky prompts.
\end{itemize}

\paragraph{Future work.}
\begin{itemize}\setlength\itemsep{1pt}
\item \textbf{Quality-weighted sampling in pretraining.} We cut the corpus at a hard rating of 1800. A softer
alternative is to keep more games and draw each one with a probability that grows with its players' ratings, as quality
filtering for language-model pretraining does, rather than scaling the loss or the learning rate by rating, which
distorts the optimization.
\item \textbf{A standard model interface.} Packing each move's fields into one integer id, and feeding the text as
pseudo-tokens whose keys and values are computed once per request, would let vLLM-style serving run the model
unchanged.
\item \textbf{Auxiliary heads}, as in Otter: predicting Stockfish's evaluation, the quality of the played move and the
result during training to shape the representations, then dropping the heads at inference.
\item \textbf{Engine-generated games}: Stockfish self-play at full strength from varied openings, for pretraining only,
since such games carry no human style.
\item \textbf{Balanced results.} In our corpus 49.1\% of games end 1-0 and 44.7\% 0-1 (\cref{sec:data}).
Sampling games and sides so that wins and losses contribute equally would give better balanced datasets.
\item \textbf{Grafting text conditioning onto a strong model.} Take one of the strong searchless models that beat ours
in \cref{sec:versus1k}, such as Maia-3, Allie 2.0 or DeepMind's searchless transformers (not an engine proxy like
Matilda), add gated cross-attention to a text encoder and train it on our described games.
\item \textbf{Cheaper and more powerful style evaluation}: playing the model against itself, one side prompted, with a
batch of a thousand games advancing one move at a time, instead of against a CPU engine.
\end{itemize}

\vspace{1.5ex}\noindent{\Large\bfseries Acknowledgments}\par\nopagebreak\vspace{1ex}\noindent
This work was supported as part of the Swiss AI Initiative by a grant from the Swiss National Supercomputing Centre
(CSCS) under project ID infra02 on Alps.

\bibliographystyle{plainnat}
\bibliography{refs}

\appendix
\section{Hyperparameters}\label{app:hparams}
\begin{table}[H]
\centering\footnotesize\setlength\tabcolsep{4pt}
\resizebox{\linewidth}{!}{\begin{tabular}{@{}lcccc@{}}
\toprule
 & pretraining & SFT & DPO (main) & DPO (full pass) \\
\midrule
initialization & random & \pt & \sft & \sft \\
text encoder & none & \makecell{EmbeddingGemma-300m\\(frozen)} & \makecell{EmbeddingGemma-300m\\(frozen)} & \makecell{EmbeddingGemma-300m\\(frozen)} \\
conditioned layers & --- & top 6 of 12 & top 6 of 12 & top 6 of 12 \\
learning rate, decoder & $5{\times}10^{-4}$ & $5{\times}10^{-7}$ & $2{\times}10^{-7}$ & $2{\times}10^{-7}$ \\
learning rate, text cross-attention & --- & $8{\times}10^{-4}$ & $10^{-4}$ & $10^{-4}$ \\
warmup, then cosine to & 5\%, 10\% & 5\%, 0 & 5\%, 0 & 5\%, 0 \\
batch per step & \makecell{128 rows of 512\\positions (32 per\\GPU, 4 GPUs)} & \makecell{up to 1,024 games\\(128 per GPU,\\8 GPUs)} &
\makecell{up to 128 games\\(1 GPU)} & \makecell{up to 2,048 games\\(256 per GPU,\\8 GPUs)} \\
steps & 1,194,960 & 28,239 & 4,366 & 13,428 \\
data & 78.32B positions & \makecell{27.08M sides\\+ top players} & 0.56M sides & 27.08M sides \\
compute & \makecell{4 GH200,\\$\approx$40\,h per pass} & 8 GH200, 3.1\,h & 1 GH200, 20\,min & 8 GH200, 2.0\,h \\
other & \makecell{illegal-mass\\penalty 0.5} & \makecell{label dropout 10\%,\\pinned no-text term} & \makecell{$\beta$ 0.5, NLL 1,\\KL 1} & \makecell{$\beta$ 0.5, NLL 1,\\KL 1} \\
\bottomrule
\end{tabular}}
\caption{Main hyperparameters of the text-conditioned pipeline.}
\label{tab:hparams}
\end{table}

\begin{table}[H]
\centering\footnotesize\setlength\tabcolsep{4pt}
\begin{tabular}{@{}lcc@{}}
\toprule
 & games-only continuation (\textsc{chsm8-player-base}) & player fine-tunes (14 models) \\
\midrule
initialization & \makecell{\pt, cross-attention replaced\\by a per-layer bias} & \textsc{chsm8-player-base} \\
trained parameters & all 166M & all 166M \\
learning rate & $5{\times}10^{-5}$ & $3{\times}10^{-5}$ \\
warmup, then cosine to & 5\%, 0 & 5\%, 0 \\
batch per step & \makecell{128 rows of 512 positions\\(32 per GPU, 4 GPUs)} & \makecell{32 rows of 512 positions\\(1 GPU)} \\
steps & 29,706 & by player, 2 passes \\
data & \makecell{26.0M games, one pass\\(1.95B positions)} & \makecell{the player's unique games (online,\\or PGN Mentor for the 4 historical\\players), 660 to 20k per player} \\
loss on & all moves & the player's own moves only \\
compute & 4 GH200, 1.1\,h & 1 GH200, at most 50\,min each \\
other & illegal-mass penalty 0.5 & illegal-mass penalty 0.5 \\
\bottomrule
\end{tabular}
\caption{Hyperparameters of the games-only branch (\cref{sec:pt-results}).}
\label{tab:hparams-games}
\end{table}

In \cref{tab:hparams}, ``text cross-attention'' is the text projection and
the gated cross-attention blocks that read the prompt; ``decoder'' is everything else. All runs use AdamW with
$(\beta_1,\beta_2)=(0.9,0.95)$, gradient clipping 1.0 and bf16.

Neither model in \cref{tab:hparams-games} has a text encoder; the
optimizer settings are as in \cref{tab:hparams}. In \pt each cross-attention block only ever saw one all-zero context
token, so its output is the same constant at every position; it is removed and that constant becomes a per-layer bias,
so training starts exactly at \pt. The alternatives tried in the search, a second self-attention in its place
(a copy of the layer's own self-attention or randomly initialized, behind a zero-initialized gate), were slower and not
better (\cref{fig:games-search}). Both models are trained with plain next-move prediction, without DPO.

\Needspace{8\baselineskip}
\begin{verified}{hparams}
\paragraph{Batching.}
Two schemes are used, chosen by how the loss is computed.
\begin{itemize}\setlength\itemsep{0pt}
\item \textbf{Greedy packing} (pretraining, games-only continuation, player fine-tunes).
Games are packed offline, in one linear pass over a shuffled game order, into rows of 512 positions: the next game is
appended if it fits; otherwise the rest of the row is padded and the game starts a new row.
No game is ever truncated, and padded positions carry no loss.
Padding is 7.7\% of the pretraining corpus.
\item \textbf{Length bucketing} (SFT and DPO).
Each game is one sequence, because its descriptions and its cross-attention context belong to that game alone.
Pools of 16,384 random games are sorted by length and cut into batches of at most 128 (SFT, main DPO) or 256
(full-pass DPO) games per GPU, under a token budget (98,304 padded positions per GPU in SFT and the main DPO, 196,608 in
the full-pass DPO); the batch order is then shuffled.
A step therefore holds fewer games when the games are long.
Games in a batch have nearly the same length, so padding is only 0.9\%; packing several games into one sequence would
save less than it costs, and was dropped.
\end{itemize}
\end{verified}

\section{Player fine-tunes}\label{app:players}
\begin{verified}{players-1}
The 14 player models of \cref{sec:pt-results} are full fine-tunes of \textsc{chsm8-player-base} on one player's own
moves (learning rate $3{\times}10^{-5}$, two passes over the player's unique games).
They are plain supervised fine-tunes: DPO on top made the fits unstable (SFT followed by DPO collapsed to 1226--1374
Elo), and supervised fine-tuning alone was enough.
\Cref{tab:player-fts} lists each of them on its held-out test games, and their strength against Stockfish at
\texttt{UCI\_Elo}~2000 (1,000 games each, 95\% CI about $\pm$19 Elo; the base model plays at 1890).
\end{verified}

\begin{table}[H]
\centering\footnotesize\setlength\aboverulesep{0pt}\setlength\belowrulesep{0pt}\renewcommand\arraystretch{1.15}\setlength\tabcolsep{3.5pt}
\resizebox{\linewidth}{!}{\begin{tabular}{@{}lrrrr|rrr|rrr|rr@{}}
\toprule
\multicolumn{5}{c}{} & \multicolumn{3}{c}{move match (\%)} & \multicolumn{3}{c}{CE (nats per move)} & \multicolumn{2}{c}{Elo} \\
\cmidrule(lr){6-8}\cmidrule(lr){9-11}\cmidrule(l){12-13}
player & FIDE rank & rating & games & positions & base & FT & $\Delta$ & base & FT & $\Delta$\,\% & FT & $\Delta$ \\
\midrule
Magnus Carlsen & 1 & 2823 & 17,561 & 1.74M & 55.4 & 56.9 & +1.5 & 2.969 & 2.915 & $-$1.8 & 1903 & +13 \\
Hikaru Nakamura & 2 & 2792 & 19,997 & 1.97M & \textbf{54.0} & \textbf{57.2} & \textbf{+3.2} & \textbf{3.039} & \textbf{2.862} & \textbf{$-$5.8} & 1850 & $-$40 \\
Fabiano Caruana & 3 & 2784 & 5,513 & 584K & 54.5 & 55.6 & +1.1 & 2.960 & 2.938 & $-$0.7 & 1902 & +12 \\
Javokhir Sindarov & 4 & 2782 & 1,371 & 135K & 55.8 & 56.7 & +0.9 & 2.959 & 2.898 & $-$2.1 & 1897 & +7 \\
Nodirbek Abdusattorov & 5 & 2780 & 5,356 & 579K & 54.4 & 54.9 & +0.5 & 2.967 & 2.995 & +0.9 & 1906 & +16 \\
Wesley So & 6 & 2770 & 5,300 & 515K & 54.7 & 56.4 & +1.7 & 2.962 & 2.969 & +0.2 & \textbf{1910} & \textbf{+20} \\
Praggnanandhaa R & 7 & 2763 & 7,467 & 729K & 54.3 & 55.0 & +0.7 & 3.016 & 3.008 & $-$0.3 & 1871 & $-$19 \\
Vincent Keymer & 8 & 2761 & 10,906 & 1.11M & 53.8 & 56.7 & +2.9 & 3.073 & 2.980 & $-$3.0 & 1887 & $-$3 \\
Wei Yi & 9 & 2758 & 2,869 & 248K & 55.2 & 55.5 & +0.3 & 3.007 & 3.015 & +0.3 & 1895 & +5 \\
Alireza Firouzja & 10 & 2757 & 19,999 & 2.07M & 55.5 & 55.9 & +0.4 & 2.962 & 2.961 & 0.0 & 1884 & $-$6 \\
\midrule
Bobby Fischer & --- & --- & 660 & 60K & 57.2 & 56.7 & $-$0.5 & 2.969 & 2.891 & $-$2.6 & 1875 & $-$15 \\
Garry Kasparov & --- & --- & 1,727 & 149K & 53.3 & 53.9 & +0.6 & 3.153 & 3.103 & $-$1.6 & 1900 & +10 \\
Anatoly Karpov & --- & --- & 3,128 & 295K & 51.9 & 54.3 & +2.4 & 3.159 & 3.044 & $-$3.6 & 1905 & +15 \\
Judit Polgar & --- & --- & 1,459 & 144K & 53.0 & 53.9 & +0.9 & 3.181 & 3.099 & $-$2.6 & 1887 & $-$3 \\
\bottomrule
\end{tabular}}
\caption{Player fine-tunes (FT) on each player's held-out test games (own moves only). Bold: the biggest change in each group.}
\label{tab:player-fts}
\end{table}

FIDE rank and standard rating in \cref{tab:player-fts} are from the FIDE top-100 list of October 2026, used for the
crawl; the four historical
players come from PGN Mentor. Games and positions are the unique training games and the positions in them (both sides,
of which the model learns the player's moves).
We collected about 20k games at most per player, in October 2026; Nakamura and Firouzja reached that cap and have
many more games online.
Each player's newest games, 2,000 or a fifth of the games if fewer, are held out and split alternately into validation
and test, so the games column counts training games only. Move match is top-1 legal move match of \textsc{chsm8-player-base} and of
the fine-tune, CE the cross-entropy in nats per move, and Elo the fine-tune's strength against Stockfish. $\Delta$ is the
change from the base model: in points of move match, in percent of the base cross-entropy, and in Elo from the base's
1890.

\begin{figure}[H]
\centering
\includegraphics[width=\linewidth]{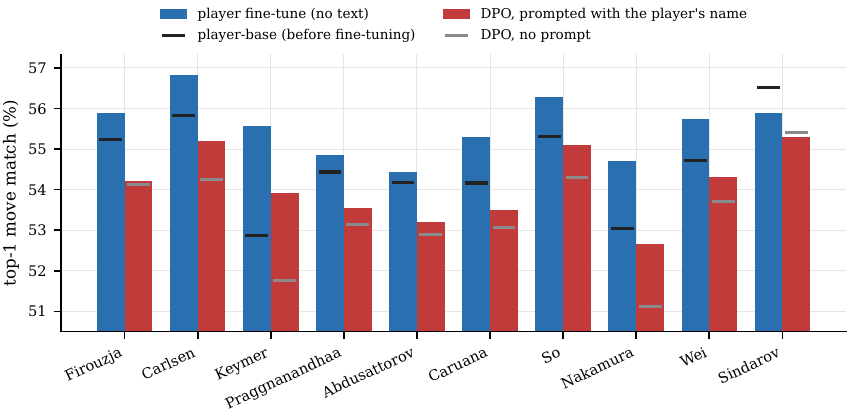}
\caption{Top-1 move match on the 1,398 held-out games of the 10 current players (ordered by number of games): each
player's own fine-tune (blue; black mark: \textsc{chsm8-player-base} before fine-tuning) and \dpo prompted with
\emph{``play like \{player\}''} (red; gray mark: \dpo with no prompt).}
\label{fig:player-ft-vs-dpo}
\end{figure}

\begin{verified}{players-2}
\paragraph{The fine-tunes mostly failed.}
Only a few players gained clearly: Nakamura ($+3.2$ points of move match), Keymer ($+2.9$), Karpov ($+2.4$) and So
($+1.7$); seven gained less than one point, and Fischer lost 0.5.
Even the largest corpus, 2.07M positions, is tiny next to the 78.32B of pretraining, and most players have 60K--730K.
That is too little to learn a personal style from: the gains are small and do not even follow the amount of data
(Firouzja, with the most positions, gained 0.4 points).
Fine-tuning on a single player would need far more of that player's games than exist online.

\paragraph{Prompted models against fine-tunes.}
For the 10 current players we also compared the text-conditioned models, prompted with \emph{``play like \{player\}''},
with the dedicated fine-tunes, on 1,398 games that none of these models saw in training.
The right name beats a wrong one for all 10 players, for both \sft and \dpo.
\sft prompted with the name reaches a slightly lower cross-entropy than the player's own fine-tune (2.957 against 2.982)
but a lower move match (54.6\% against 55.5\%): a single prompted model spreads its probability toward each player
about as well as a dedicated fine-tune, while the fine-tune picks the player's exact move more often
(\cref{fig:player-ft-vs-dpo}).
\end{verified}

\section{A labeled example}\label{app:examples}
\begin{verified}{examples}
One side of a held-out 3-minute game, described by all three labelers: White, rated 2277, against a 1891 opponent in a
Sicilian Defense, Nyezhmetdinov--Rossolimo Attack (\texttt{1.e4 c5 2.Nf3 Nc6 3.Bb5 Qb6 4.Bxc6 Qxc6 5.O-O Qxe4
6.Re1\ldots}), won in 25 moves.
White gives up the e4 pawn early for development and then attacks the black king.
The labeler's input is the move list and this card, abridged:

\begin{quote}\small\ttfamily
development: brisk (3rd piece out at ply 17); king: kingside on ply 9;\\
captures: 10 (0.40/move, very bloody); checks given: 6 (0.24/move);\\
sacrifices: 0, total invested 0.0 pawns; worst material deficit endured: 2.2 pawns;\\
central presence: controls it; pawn moves: 0.12/move; pawn pushes at the enemy king: 0;\\
pressure on black's king ring: 2.32 attackers/ply --- real; queens: stayed on the board
\end{quote}

\end{verified}
\begin{table}[H]
\centering\footnotesize\setlength\tabcolsep{4pt}
\begin{tabular}{@{}lp{0.27\linewidth}p{0.27\linewidth}p{0.27\linewidth}@{}}
\toprule
band & Gemma-4-31B (teacher) & Gemma-3-1B (distilled, clean) & Apertus-v1.5-70B \\
\midrule
terse & give up a pawn to blitz the king & give up a pawn for a fast attack & trade pieces and keep pushing pawns \\
terse & hunt the queen then clean up the board & hunt the king with everything & attack, sacrifice, don't hold back \\
\midrule
short & ignore early material losses to create an overwhelming attack against the exposed king. &
play recklessly by offering material to gain a devastating lead in development and initiative. &
sacrifice calmly and salt kingside with pawns while trading everything else \\
\midrule
medium & play with reckless aggression by trading pawns for rapid development and using the knights to dismantle the
opponent's position quickly. &
ignore the early loss of a central pawn to launch a chaotic and bloody offensive that keeps the enemy king running. &
play fast and loose with material, keep queens on the board, and pile pawns on the king after sacrificing pieces \\
\midrule
long & sacrifice an early pawn for a lead in development and then launch a chaotic offensive that captures almost
everything in sight until the king is trapped in the center of the board. &
be completely fearless with material and prioritize speed over safety so you can swarm the opponent with pieces while
they are still struggling to coordinate their defense in a total collapse. &
ignore safe positions and keep sacrificing pieces to crash the kingside with pawns, trade relentlessly, and never stop
pressing forward---this blitz pace rewards aggression over caution \\
\bottomrule
\end{tabular}
\caption{The five descriptions of the same side from each labeler. Gemma and its distilled student both describe the
pawn given up for development and the attack on the king. Apertus asks for piece sacrifices and a kingside pawn storm,
which the card rules out (no sacrifice, no pawn push at the enemy king).}
\label{tab:example}
\end{table}

\section{Open release and reproducibility}\label{app:artefacts}
\sloppy
\begin{verified}{artefacts}
The project is fully open source.
Everything needed to reproduce each stage, or to build on any intermediate result, is public.
You are welcome to reproduce our results, build on the models, and use the datasets and labels in your own work.

\paragraph{Weights.}
\begin{itemize}\setlength\itemsep{0pt}
\item \hf{chsm8-pt}: pretrained model (1837 Elo).
\item \hf{chsm8-player-base}: games-only continuation (1890 Elo), and 14 \texttt{chsm8-player-*} fine-tunes:
Magnus Carlsen, Hikaru Nakamura, Fabiano Caruana, Wei Yi, Wesley So, Alireza Firouzja, Vincent Keymer, Nodirbek
Abdusattorov, Javokhir Sindarov and Praggnanandhaa R (current players), and Bobby Fischer, Garry Kasparov, Anatoly Karpov
and Judit Polgar (historical).
\item \hf{chsm8-labeler-gemma3-1b}: the distilled labeler (clean weights on \texttt{main}).
\item \hf{chsm8-sft}: the style-conditioned SFT model.
\item \hf{chsm8}: the final model (the main 20-minute DPO run, \texttt{model.pt}), and the full-pass DPO model with its checkpoints in
\texttt{full\_pass/}; \texttt{onnx/} holds the browser export (8-bit weights, fp32 compute).
\end{itemize}

\paragraph{Datasets.}
\begin{itemize}\setlength\itemsep{0pt}
\item \hfd{chsm8-pre-training}: 1.07B filtered Lichess games.
\item \hfd{chsm8-game-labels}: all labels written by the teacher LLMs, served on the SwissAI platform
(\cref{sec:teachers}): Gemma-4-31B (3.02M sets, 4.6\% of bands contradicted) and Apertus-v1.5-70B (4.75M sets, thinking
on and off, 16.5\% contradicted), each raw and after our repair passes; plus the same Apertus sides relabeled from scratch
by the clean distilled labeler (0.40\% of bands contradicted).
The different contradiction rates, and why only Gemma was used as a teacher, are in \cref{sec:teachers,sec:label-signal}.
\item \hfd{chsm8-distillation}: the exact distillation set of the clean labeler: the 2.40M Gemma-4-31B label sets in
which no band contradicts the card, with the full model input (moves and feature card), plus unfiltered validation
and test splits.
\item \hfd{chsm8-post-training}: the 30.1M described game sides SFT and DPO were trained on, with the same train,
validation and test splits; every label set records whether it was written by Gemma-4-31B or by the distilled labeler,
and which bands pass the card check.
\item \hfd{chsm8-top-players}: the top-player games used for player prompts and the player fine-tunes, with their
held-out split.
\end{itemize}

\paragraph{Prompts and data scripts.}
\begin{itemize}\setlength\itemsep{0pt}
\item \textbf{Labelling}: the teacher prompt, the feature-card builder and the full labeling, repair and export pipeline
ship with \hfd{chsm8-post-training}, \hfd{chsm8-game-labels} and \hfd{chsm8-distillation} (\texttt{scripts/}); \texttt{build\_prompts.py}
rebuilds the exact labeler input for any described game side.
\item \textbf{Pretraining data}: \hfd{chsm8-pre-training} includes the scripts that filtered the Lichess dumps and
those that convert the corpus into packed training records.
\item \textbf{Top players}: \hfd{chsm8-top-players} includes the crawler and the 49 ``play like \{player\}'' prompt
templates with their name aliases.
\item \textbf{Evaluation prompts}: the full prompt set (strong, weak and style prompts) is in \hf{chsm8-sft}
(\texttt{eval/prompts.json}); the four main prompts are listed in \cref{sec:eval}.
\item \textbf{Configurations}: all hyperparameters are in \cref{app:hparams}.
\end{itemize}

\paragraph{Move tokenizer.}
\hf{chsm8-tokenizer} turns games into exactly the move tokens \chsm was trained on, so games from other sources can be
used with the models or tokenized the same way.
It accepts SAN, LAN, UCI and full PGN (move numbers, comments, variations and annotations), mixed freely within one game,
and checks every move for legality.
The core is in Rust on \texttt{shakmaty}, with Python and WebAssembly bindings, and tokenizes about 2M moves per second.

\paragraph{Code and demos.}
The inference code, the move tokenizer (Rust, with Python and WebAssembly bindings), the ONNX export and two
playable demos (in the browser, and on a local GPU) are at \url{https://github.com/nov3o/llm_chess-demo}.
The browser demo and the project page are live at \url{https://llmchess-demo-web.vercel.app}.
The training and evaluation code, the search logs behind every figure and the scripts that rebuild the figures and
tables will be released later.

\paragraph{Licenses.}
Our models and code are released under MIT.
The labeler is a fine-tune of Gemma~3 and is therefore distributed under the Gemma Terms of Use.
Both teacher LLMs, Gemma-4-31B and Apertus-v1.5-70B, are released under Apache-2.0, and so are the datasets
that contain their labels or labels derived from them (\hfd{chsm8-game-labels}, \hfd{chsm8-distillation} and \hfd{chsm8-post-training}).
The Lichess games are CC0.
The prompt encoder, EmbeddingGemma-300m, is not redistributed: it is downloaded at run time under the Gemma Terms of Use.
\end{verified}

\section{Bradley--Terry strength}\label{app:bt}
\begin{wraptable}{r}{0.61\linewidth}
\vspace{-12pt}
\centering\footnotesize\setlength\tabcolsep{2.4pt}
\begin{tabular}{@{}rlr@{\,}lrrl@{}}
\toprule
\# & model & \multicolumn{2}{c}{BT (95\% CI)} & games & Elo & Elo rank \\
\midrule
1 & \grpsearch\,Matilda & 2674 & $\pm$\,39 & 4,000 & 3009 & \makebox[0.9em][r]{1}\,\rankeq \\
2 & \grpadv\,Searchless 136M & 2431 & $\pm$\,21 & 3,999 & 2584 & \makebox[0.9em][r]{2}\,\rankeq \\
3 & \grpadv\,Searchless 270M & 2418 & $\pm$\,21 & 3,982 & 2505 & \makebox[0.9em][r]{3}\,\rankeq \\
4 & \grpadv\,Searchless 9M & 2321 & $\pm$\,17 & 3,996 & 2403 & \makebox[0.9em][r]{4}\,\rankeq \\
5 & \grpadv\,QUEEN (PAWN-8) & 2291 & $\pm$\,34 & 800 & 2354 & \makebox[0.9em][r]{5}\,\rankeq \\
6 & \grpadv\,Maia-3 79M & 2271 & $\pm$\,15 & 3,997 & 2271 & \makebox[0.9em][r]{6}\,\rankeq \\
7 & \grpadv\,Allie 2.0 & 2250 & $\pm$\,15 & 4,000 & 2232 & \makebox[0.9em][r]{7}\,\rankeq \\
8 & \grpsup\,\textbf{\textsc{player-carlsen}} & 2022 & $\pm$\,10 & 16,199 & 1903 & \makebox[0.9em][r]{11}\,\rankup{3} \\
9 & \grpsup\,\textbf{\textsc{player-base}} & 2006 & $\pm$\,10 & 16,200 & 1890 & \makebox[0.9em][r]{12}\,\rankup{3} \\
10 & \grpsearch\,Stockfish (anchor) & 2000 & (fixed) & 18,146 & --- & \makebox[0.9em][r]{10}\,\rankeq \\
11 & \grpadv\,Otter 15M & 1972 & $\pm$\,12 & 3,999 & 2021 & \makebox[0.9em][r]{8}\,\rankdown{3} \\
12 & \grpsearch\,ChessLFM with search & 1969 & $\pm$\,12 & 3,985 & 2001 & \makebox[0.9em][r]{9}\,\rankdown{3} \\
13 & \grpsup\,\textbf{\pt} & 1953 & $\pm$\,9 & 17,199 & 1837 & \makebox[0.9em][r]{14}\,\rankup{1} \\
14 & \grpsup\,Maia-2 blitz & 1814 & $\pm$\,13 & 4,198 & 1657 & \makebox[0.9em][r]{19}\,\rankup{5} \\
15 & \grpsup\,Chess-GPT (Karvonen) & 1786 & $\pm$\,14 & 4,199 & 1873 & \makebox[0.9em][r]{13}\,\rankdown{2} \\
16 & \grpsup\,chess-gpt2-uci & 1691 & $\pm$\,16 & 4,197 & 1784 & \makebox[0.9em][r]{15}\,\rankdown{1} \\
17 & \grpsup\,ChessFormer (Transcendence) & 1642 & $\pm$\,17 & 4,200 & 1733 & \makebox[0.9em][r]{16}\,\rankdown{1} \\
18 & \grpsup\,ChessGPT-base 2.8B & 1619 & $\pm$\,18 & 4,199 & 1731 & \makebox[0.9em][r]{17}\,\rankdown{1} \\
19 & \grpsup\,ChessLFM & 1595 & $\pm$\,20 & 3,997 & 1687 & \makebox[0.9em][r]{18}\,\rankdown{1} \\
20 & \grpadv\,C1-4B & 1164 & $\pm$\,59 & 4,000 & 1112 & \makebox[0.9em][r]{20}\,\rankeq \\
\bottomrule
\end{tabular}

\caption{Bradley--Terry strength (BT) with its 95\% interval from all 64,746 games, the games each model played, and the
Elo it gets from Stockfish alone (\cref{tab:versus1k}). Elo rank: position by that Elo; triangles: places gained (green) or lost (red) under BT.
Ours in bold. \grpsup\ supervised move prediction, \grpadv\ strong recent models without search, \grpsearch\ search.}
\label{tab:bt}
\vspace{-8pt}
\end{wraptable}

The Elo numbers in \cref{tab:versus1k} each come from a single opponent: a model's score $S$ against Stockfish at
\texttt{UCI\_Elo}~2000 gives $2000+400\log_{10}\!\big(S/(1-S)\big)$.
That scale is reproducible, but it measures how well a model handles one engine that plays strong moves and inserts
deliberate mistakes, and the head-to-head matches show it is not transitive: \textsc{chsm8-player-base} rates 1890
against Stockfish, below ChessLFM with search (2001) and Otter (2021), yet it wins both matches, 57.5\% and 56.1\%.

A Bradley--Terry model \citep{bradley1952rank} instead fits one strength $r_i$ per player to all games at once.
On the Elo scale, player $i$ scores against player $j$ with expectation
\[
\sigma_{ij}=\frac{1}{1+10^{-(r_i-r_j)/400}} .
\]
For each pairing with $N_{ij}$ games and score $S_{ij}=W_{ij}+D_{ij}/2$ (a draw counts as half a win for each side),
the log-likelihood is
\[
\ell(r)=\sum_{(i,j)} S_{ij}\log\sigma_{ij}+\left(N_{ij}-S_{ij}\right)\log\left(1-\sigma_{ij}\right).
\]
It depends only on differences, so we fix Stockfish at $r=2000$.
With $c=\ln 10/400$, the gradient is $\partial\ell/\partial r_i=c\sum_j\left(S_{ij}-N_{ij}\sigma_{ij}\right)$ and each
pairing adds $c^2N_{ij}\sigma_{ij}(1-\sigma_{ij})$ to the information matrix.
Newton's method converges in 9 steps; the intervals are $\pm1.96$ standard errors from the inverse information.

The fit uses every match we played: 72 pairings among 20 players.
Every model played Stockfish (1,000 games; QUEEN 200), our three models played every public model (1,000 games;
QUEEN 200), and \pt also played the 200-game matches of \cref{tab:versus}.
Public models never played each other, so they are linked only through Stockfish and our models.

Our models rise about 115 points, and the two games-trained ones pass Stockfish, Otter and ChessLFM with search;
models that score well against the limited engine, such as ChessLFM with search, Matilda and the searchless transformers, fall.
The ranking at the top does not change.
One number per model cannot absorb the remaining intransitivity, so the intervals are optimistic, and the ranking is
the more reliable part of \cref{tab:bt}.

\end{document}